\documentclass[sigconf, nonacm]{acmart}

\usepackage{pvldb}

\renewcommand\vldbdoi{XX.XX/XXX.XX}
\renewcommand\vldbpages{XXX-XXX}
\renewcommand\vldbavailabilityurl{}

\newcommand{\ours}{\textsc{Coal-SQL}\xspace}

\newcommand{\solveNone}{\textsc{solve\_none}\xspace}
\newcommand{\skel}{\sigma}                       
\newcommand{\emb}{\phi}                           

\usepackage{xspace}
\usepackage{hyperref}           
\usepackage{url}                
\usepackage{booktabs}           
\usepackage{amsfonts}           
\usepackage{amsmath}            
\usepackage{mathtools}          
\usepackage{nicefrac}           
\usepackage{microtype}          
\usepackage{xcolor}             
\usepackage{multirow}           
\usepackage{graphicx}           
\usepackage{enumitem}           
\usepackage{algorithm}          
\usepackage{algpseudocode}      

\usepackage{xcolor}

\newcommand{\masktoken}{%
  \begingroup
  \setlength{\fboxsep}{1pt}%
  \colorbox{red!10}{%
    \textcolor{red!65!black}{%
      \texttt{\textless MASK\textgreater}%
    }%
  }%
  \endgroup
}

\newcommand{\numtoken}{%
  \begingroup
  \setlength{\fboxsep}{1pt}%
  \colorbox{red!10}{%
    \textcolor{red!65!black}{%
      \texttt{\textless NUM\textgreater}%
    }%
  }%
  \endgroup
}

\begin{document}
\title{COAL-SQL: Coverage-Guided Augmentation and Failure-Driven Learning for Text-to-SQL Post-Training}

\author{Qifeng Cai}
\authornote{Equal contribution.}
\affiliation{%
  \institution{Peking University}
  \city{Beijing}
  \country{China}
}
\email{qifeng.cai04@gmail.com}

\author{Xuanguang Pan}
\authornotemark[1]
\affiliation{%
  \institution{Beihang University}
  \city{Beijing}
  \country{China}
}
\email{panxg03@gamil.com}

\author{Hao Liang}
\authornotemark[1]
\authornote{Project leader.}
\affiliation{%
  \institution{Peking University}
  \city{Beijing}
  \country{China}
}
\email{hao.liang@pku.edu.cn}

\author{Chang Xu}
\affiliation{%
  \institution{Peking University}
  \city{Beijing}
  \country{China}
}
\email{cxu25@pku.edu.cn}

\author{Wentao Zhang}
\authornote{Corresponding author.}
\affiliation{%
  \institution{Peking University}
  \city{Beijing}
  \country{China}
}
\email{wentao.zhang@pku.edu.cn}

\begin{abstract}
Text-to-SQL translates natural-language questions into executable SQL
queries, yet open-source large language models (LLMs) still require
task-specific post-training for complex SQL generation.
Effective post-training requires both high-quality data that cover the
capabilities required by the target task and a learning strategy that
enables the model to acquire them. Existing Text-to-SQL datasets
provide valuable supervision, but their structural coverage remains
incomplete. Existing augmentation methods often expand the corpus
without identifying what is missing from the available data. A more
efficient strategy is to identify these structural gaps and complement
the existing dataset in a targeted manner. Meanwhile, relying solely
on supervised fine-tuning (SFT) or reinforcement learning (RL) can
make it difficult to dynamically strengthen the weaknesses exposed by
the model during training.
We propose \ours, a unified post-training framework that combines
\textbf{Co}verage-Guided \textbf{A}ugmentation with Failure-Driven
\textbf{L}earning. Coverage-Guided Augmentation uses greedy selection
to identify SQL structures missing from the existing dataset and
constructs complementary examples, thereby improving structural
coverage. Failure-Driven Learning retains GRPO as the main objective
and provides targeted supervision for examples the model fails to
solve. At the step level, it performs SFT on strong-LLM-generated
reasoning traces for Solve-None examples. At the epoch level, it
retrieves structurally related examples for accumulated failures to
construct additional practice, helping the model acquire the
corresponding SQL capabilities. With only 12.6K distinct post-training
examples, \ours achieves 64.9\% execution accuracy on the BIRD
development set and outperforms comparable-scale baselines,
demonstrating its effectiveness. The code is available at
\url{https://github.com/TechNomad-ds/COAL-SQL}.
\end{abstract}

\maketitle

\vldbtopmatter

\section{Introduction}
\label{sec:introduction}

Relational databases offer a robust foundation for managing structured data, yet querying them typically demands in-depth knowledge of Structured Query Language (SQL), which poses a significant barrier for non-experts. Text-to-SQL aims to address this challenge by automatically translating natural-language (NL) questions into executable SQL queries~\citep{yu2018spider,li2023bird,liu2024nl2sqlsurvey}. While large language models (LLMs) have considerably advanced the state of the art in Text-to-SQL, open-source models continue to struggle with intricate query structures and domain-specific database schemas. Consequently, specific post-training remains a crucial step for enhancing the Text-to-SQL performance of these models.

Two key factors determine the effectiveness of post-training: (1) \textit{whether the training data provides the model with sufficient coverage of SQL capabilities}, and (2) \textit{whether the learning strategy enables the model to effectively acquire those capabilities}. The first factor defines the upper bound of what the model can potentially learn, while the second governs whether the available supervision is successfully translated into actual model performance.

\textbf{From the data perspective}, existing Text-to-SQL datasets are typically derived from publicly available resources or constructed through manual annotation. These datasets provide high-quality and valuable supervisory signals; however, their construction demands substantial database expertise and annotation efforts, rendering further expansion costly~\citep{guo2018question,wu2021data}. Moreover, their coverage remains inherently incomplete, as finite datasets can only represent a subset of the diverse SQL paradigms, thereby limiting model generalization to under-represented query types.

In response, recent research has shifted toward automatic synthesis methods to augment training corpora~\citep{li2025omnisql,cai2025text2sqlflow,yang2024sense}. For instance, OmniSQL constructs approximately 2.5 million training examples, significantly expanding both the scale and diversity of available supervisory signals~\citep{li2025omnisql}. The effectiveness of such large-scale synthesis is indisputable. Nevertheless, when a valuable dataset already exists, relying primarily on corpus size may not be the most data-efficient scaling strategy. Existing data may already encompass numerous useful capabilities, with deficiencies present only in specific areas. Expanding the corpus indiscriminately without analyzing current coverage proves inefficient.

\begin{figure}[t]
\centering
\includegraphics[width=0.9\linewidth]
{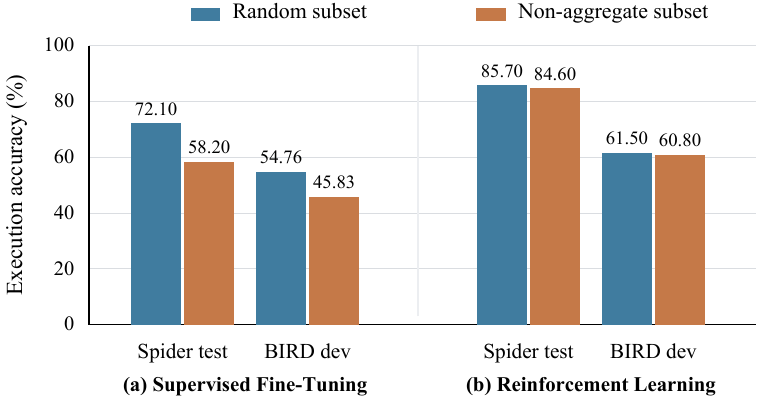}
\caption{Performance of equally sized training subsets with different
SQL structural coverage.}
\label{fig:seed-coverage-motivation}
\end{figure}

Therefore, the core issue is not how to replace existing datasets with larger corpora, but how to identify what aspects have already been covered by the available data and selectively supplement the missing parts. This necessitates a practical foundation for characterizing coverage. The SQL structure offers such a foundation, as SQL queries are highly compositional, involving diverse combinations of operators such as joins, aggregations, nested queries, and set operations~\citep{yu2018spider,zhu2025sacsql}. The structural composition of a dataset thus reflects which query patterns are represented and which remain insufficiently covered. As illustrated in Figure~\ref{fig:seed-coverage-motivation}, we construct two subsets from the BIRD training set, each containing 3,000 samples. One subset is obtained via random sampling, while the other excludes queries involving aggregation operations. With all other training configurations held identical, the model trained on the aggregation-excluded subset underperforms in both supervised fine-tuning and reinforcement learning settings, with more pronounced degradation on aggregation-related questions. This underscores the impact of the SQL structures covered by the data on training efficacy. Indiscriminate synthesis may repeatedly reinforce already well-represented structures while failing to adequately cover under-represented compositions.

\textbf{From the learning perspective}, broader structural coverage determines
which SQL capabilities the model encounters, but exposure alone does
not guarantee that those capabilities are acquired. An effective
learning strategy should identify and strengthen the weaknesses
exposed by the evolving model during training. Relying solely on supervised fine-tuning (SFT) or reinforcement learning (RL) makes this difficult for different reasons. SFT typically learns
from supervision constructed before optimization and therefore cannot
adapt its emphasis to newly observed failures. RL can learn from the
model's generated outcomes, but its effectiveness depends on whether
the sampled behaviors provide useful learning signals.

In Text-to-SQL, execution rewards provide a direct measure of SQL
correctness~\citep{kulkarni2025reinforcing,yao2025arctic}. Group
Relative Policy Optimization (GRPO), for example, samples a group of
outputs for each question and updates the policy according to their
relative rewards~\citep{shao2024deepseekmath}. When a group contains
both correct and incorrect execution rollouts, their reward
contrast provides a direct signal for reinforcing successful SQL
behavior. For some examples, however, none of the sampled rollouts is
execution-correct. We refer to an example as \emph{Solve-None} when
all of its rollouts are execution-incorrect under the current policy
and rollout budget. Auxiliary rewards may still distinguish properties
such as output format, but the group contains no correct SQL behavior
that indicates how the example should be solved.
Under the initial policy, we find that 15.1\% of our training examples are
Solve-None. These failures are concentrated on complex structures
involving nested queries, multi-table joins, and combinations of
aggregation operations. Unlike structural gaps identified from the
static training corpus, Solve-None examples reveal capability gaps
specific to the current policy. Moreover, because the policy continues
to evolve, the set of unresolved examples also changes throughout
training.

These dynamically exposed failures provide a natural basis for
targeted supervision. For each Solve-None example, a strong LLM can
construct a corrective chain-of-thought (CoT) reasoning trace,
supplying the positive SQL behavior absent from its rollouts. However,
supervision tied only to the failed instance may not generalize
sufficiently to other examples requiring the same SQL constructs.
Structurally related practice is therefore also needed to reinforce
the underlying query organization and error-prone operations. Direct
correction addresses the current failure, while related practice
strengthens the SQL capability responsible for that failure. Both
forms of supervision should be constructed from failures observed
during training rather than fixed entirely in advance.

To jointly address the data and learning challenges, we propose
\ours, a unified Text-to-SQL post-training framework that combines
\textbf{Co}verage-Guided \textbf{A}ugmentation with Failure-Driven
\textbf{L}earning. Coverage-Guided Augmentation determines which SQL
structures should be introduced before policy optimization, while
Failure-Driven Learning dynamically strengthens the capabilities that
the evolving policy still fails to acquire.

\paragraph{Coverage-Guided Augmentation.}
This expands the seed set by leveraging its existing structural coverage. It abstracts SQL queries into schema-independent skeletons and applies a greedy metric \(K\)-center selection strategy, fixing the seed skeletons as the initial centers. This selection mechanism prioritizes structures that complement the seed set while preserving diversity among the selected candidates. Each chosen skeleton is subsequently instantiated on a target training database and paired with a generated natural language question, thereby filling structural gaps in the original dataset and enhancing its overall structural coverage.

\paragraph{Failure-Driven Learning.}
This retains GRPO as the primary learning objective
and provides targeted supervision for failures exposed during
optimization. Step-level expansion collects newly observed Solve-None
examples and uses a strong LLM to construct corrective CoT reasoning
traces for batched SFT. Epoch-level expansion further constructs
structure-related practice from failures accumulated during an epoch.
It retrieves candidate examples according to their gold SQL skeletons,
diagnoses the primary SQL error in each failed rollout, and uses the
diagnosed error type to refine the skeleton-based ranking when
appropriate. CoT reasoning traces are then constructed for the
selected practice examples. Step-level expansion directly corrects
failed instances, while epoch-level expansion strengthens the
corresponding SQL capabilities.

Our contributions are summarized as follows:

\begin{enumerate}[leftmargin=1.3em,topsep=3pt,itemsep=2pt]

\item We propose \textbf{Coverage-Guided Augmentation}, which identifies structural gaps in an existing Text-to-SQL seed set and constructs complementary examples on target training databases. This approach improves structural coverage while preserving task relevance.

\item We propose \textbf{Failure-Driven Learning}, which retains GRPO as the primary learning objective and converts dynamically observed Solve-None failures into two forms of targeted supervision: step-level corrective CoT reasoning traces and epoch-level structure-related practice.

\item Together, these two components form \ours. With approximately 12.6K distinct post-training examples, \ours achieves 64.9\% execution accuracy on the BIRD development set. Cross-benchmark evaluation and ablation studies further demonstrate the effectiveness of the two proposed components.

\end{enumerate}

\section{Related Work}
\label{sec:related-work}

\subsection{Text-to-SQL Data Construction}

Early Text-to-SQL research primarily relied on human-annotated
datasets. WikiSQL focuses on single-table queries, whereas Spider
introduces complex cross-domain databases and compositional SQL
structures~\citep{zhong2017seq2sql,yu2018spider}. Subsequent
benchmarks extend the task to conversational interaction, larger and
more realistic databases, realistic data values, and external
knowledge~\citep{yu2019sparc,yu2019cosql,li2023bird}. Although these
datasets provide high-quality supervision and evaluation resources,
their construction requires substantial database expertise and
annotation effort, limiting further expansion in scale and structural
diversity. To reduce annotation costs, earlier work synthesizes Text-to-SQL
examples using templates, grammars, abstract syntax trees,
SQL-to-question generation, cycle consistency, and
self-play~\citep{guo2018question,wu2021data,zhong2020grounded,
liu2022augmenting}. Recent LLM-based pipelines support more flexible
question and SQL generation, execution-based filtering, CoT reasoning
trace construction, and preference-data
synthesis~\citep{yang2024sense,duan2025dsqg,cai2025text2sqlflow, li2025omnisql,zhu2025sacsql}. These studies improve
the scale, validity, diversity, and difficulty distribution of
synthetic data.

\subsection{Text-to-SQL Post-Training}

Text-to-SQL models are commonly adapted through SFT on question--SQL pairs, with recent approaches
further incorporating CoT reasoning traces to
improve complex SQL generation~\citep{he2025starsql}. At inference
time, execution-guided decoding uses execution results to reject
invalid partial programs~\citep{wang2018executionguided}. Other
methods improve SQL generation through constrained decoding, task
decomposition, and iterative correction
~\citep{scholak2021picard,pourreza2023dinsql}.

Recent work incorporates execution feedback into Text-to-SQL
optimization through preference-data construction, query-level
execution rewards, outcome- or step-level reward modeling, and
multi-step database interaction
~\citep{yang2024sense,kulkarni2025reinforcing,
yao2025arctic,weng2025graphreward,dai2025reexsql,hua2026sqltrail}.
These methods enable models to learn from the execution outcomes of
generated SQL queries. However, in group-relative optimization such as
GRPO, when all rollouts sampled for an example are
execution-incorrect, the group contains no successful SQL behavior to
reinforce. Related reasoning methods address similar limitations by
supplementing reinforcement learning with successful reasoning traces:
ReFT uses supervised warm-up before online reinforcement
learning~\citep{luong2024reft}, LUFFY incorporates off-policy
reasoning traces~\citep{yan2025luffy}, and ReLIFT interleaves
reinforcement learning with supervised updates on difficult questions
identified during training~\citep{ma2025relift}. These studies
demonstrate the value of additional positive supervision when
reward-based exploration is insufficient. Nevertheless, dynamically
constructing Text-to-SQL-specific supervision and related practice
from execution failures during training remains underexplored.

\begin{figure*}[t]
    \centering
    \includegraphics[width=\linewidth]{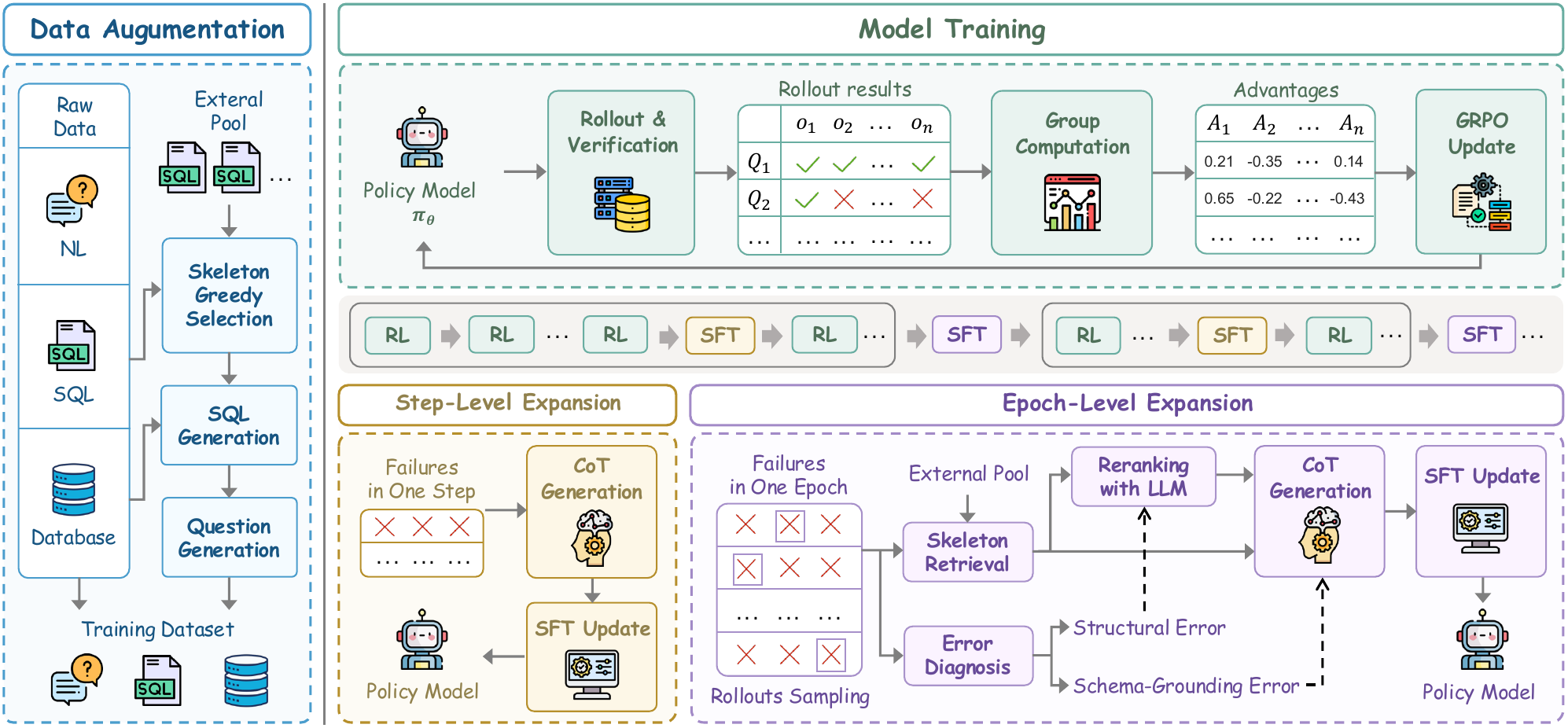}
\caption{Overview of \ours. The left panel illustrates
Coverage-Guided Augmentation, which identifies structural gaps in the
seed set and constructs complementary data. The right panel
illustrates Failure-Driven Learning, which supplements GRPO with
step-level expansion to correct Solve-None examples and epoch-level
expansion to construct structure-related practice from accumulated
failures.}
    \label{fig:framework}
\end{figure*}

\section{Preliminaries}
\label{sec:prelim}

\paragraph{Text-to-SQL formulation.}

A Text-to-SQL example is represented as a triple
$(q_i,s_i,\mathrm{db}_i)$, where $q_i$ is the NL question, $s_i$ is its gold SQL query, and $\mathrm{db}_i$ is
the associated database. The training set is
\begin{equation}
\label{eq:training-set}
\mathcal{D}
=
\left\{
(q_i,s_i,\mathrm{db}_i)
\right\}_{i=1}^{|\mathcal{D}|}.
\end{equation}
Let \(S_i = \operatorname{schema}(\mathrm{db}_i)\) denote the relational schema of \(\mathrm{db}_i\), represented as the \texttt{CREATE TABLE} statements of all tables, and let \(C_i = \operatorname{SampleContent}(\mathrm{db}_i)\) denote the sampled database content provided to the model, consisting of three sampled rows per table formatted as \texttt{INSERT} statements. We construct the model input as
\begin{equation}
\label{eq:prompt-construction}
P_i = \operatorname{Serialize}\left(S_i, C_i, q_i\right).
\end{equation}

Given \(P_i\), the LLM policy \(\pi_\theta\) performs reasoning to generate an output \(o_i \sim \pi_\theta(\cdot \mid P_i)\), from which we extract the predicted SQL query \(\hat{s}_i = \operatorname{extract}(o_i)\). For any SQL query
$s$ and database $\mathrm{db}$, we define its normalized execution
result as
\begin{equation}
\label{eq:execution-result}
\operatorname{Ans}(s,\mathrm{db})
=
\begin{cases}
\operatorname{Norm}\!\left(
\operatorname{Exec}(s,\mathrm{db})
\right),
& \text{if $s$ executes successfully},\\
\bot,
& \text{otherwise}.
\end{cases}
\end{equation}
Two SQL queries $s$ and $s'$ are execution-equivalent on
$\mathrm{db}$, denoted by $s\equiv_{\mathrm{db}}s'$, if they execute
successfully and produce the same normalized result:
\begin{equation}
\label{eq:execution-equivalence}
s\equiv_{\mathrm{db}}s'
\iff
\operatorname{Ans}(s,\mathrm{db})\neq\bot
\;\land\;
\operatorname{Ans}(s,\mathrm{db})
=
\operatorname{Ans}(s',\mathrm{db}).
\end{equation}
Result normalization follows the benchmark evaluation protocol.
Row multiplicities are preserved, row order is ignored unless the
query explicitly specifies an ordering, and numerical and
\texttt{NULL} values are normalized consistently.

\paragraph{SQL skeletons.}

Following prior skeleton-based Text-to-SQL methods
\citep{gao2024dailsql,li2023resdsql}, we use SQL skeletons to represent
database-independent query structures. Given a parsed SQL query $s$,
the skeletonization function $\skel(\cdot)$ replaces database-specific
identifiers, including table names, column names, and aliases, with a
generic mask token, while replacing literals with type-specific
placeholders. SQL operators, clause organization, join-clause
organization, set operations, and nesting relationships are preserved.
Superficial variations such as keyword casing and alias naming are
also canonicalized.

For example, the query
\texttt{SELECT name FROM employee WHERE age > 18 ORDER BY name}
is canonicalized into the SQL skeleton
\texttt{SELECT }\masktoken\texttt{ FROM }\masktoken
\texttt{ WHERE }\masktoken\texttt{ > }\numtoken
\texttt{ ORDER BY }\masktoken.
Queries over different databases can therefore share the same
skeleton when they instantiate the same clause and operator
composition.
This abstraction intentionally ignores the identities of individual
tables and columns.

\paragraph{CoT Reasoning Trace Generation.}
For questions that the weak model answers incorrectly, we leverage a strong LLM to generate a chain-of-thought (CoT) reasoning trace $\tau_i$ and subsequently fine-tune the weak model on these traces (i.e., supervised fine-tuning, SFT) to improve its capability.
Following prior work on explicit reasoning supervision~\citep{wei2022cot,he2025starsql}, the strong model receives the problem $P_i$ together with the gold SQL query $s_i$ and produces the trace
\begin{equation}
\label{eq:cot-trace-construction}
\tau_i = M_{\mathrm{strong}}(P_i, s_i).
\end{equation}
The gold SQL query is provided only as a reference; the prompt instructs the model to reason naturally from $P_i$ itself.
Let $s_i^\tau = \operatorname{extract}(\tau_i)$ denote the final predicted SQL query extracted from the generated trace.
We retain $\tau_i$ only if the extracted query is execution-equivalent to the gold SQL on the associated database, i.e., $s_i^\tau \equiv_{\mathrm{db}_i} s_i$.
If not, the trace is discarded, and we sample multiple times for such erroneous cases until a valid $\tau_i$ is obtained.
This procedure verifies only the final SQL query and does not check the intermediate reasoning steps.

\paragraph{Supervised Fine-Tuning.}
The accepted CoT reasoning traces form the supervised dataset $\mathcal{T}=\{(P_i,\tau_i)\}_{i=1}^{|\mathcal{T}|}$. We optimize the policy with the standard
autoregressive objective:
\begin{equation}
\label{eq:sft-loss}
\mathcal{L}_{\mathrm{SFT}}(\theta)
=
-
\mathbb{E}_{(P_i,\tau_i)\sim\mathcal{T}}
\left[
\sum_{j=1}^{|\tau_i|}
\log
\pi_\theta
\left(
\tau_{i,j}
\mid
P_i,\tau_{i,<j}
\right)
\right].
\end{equation}

\paragraph{Group Relative Policy Optimization.}

Group Relative Policy Optimization (GRPO)
\citep{shao2024deepseekmath} optimizes the policy using relative
rewards within a group of sampled outputs, without introducing a
separate value model, while retaining the clipped surrogate
optimization used by PPO~\citep{schulman2017ppo}. For prompt $P_i$,
GRPO samples a group of $N$ rollouts
$\{o_i^{(k)}\}_{k=1}^{N}$ from
$\pi_{\theta_{\mathrm{old}}}(\cdot\mid P_i)$. Let
$R_i^{(k)}=R(o_i^{(k)},s_i,\mathrm{db}_i)$ denote the reward of the
$k$-th rollout. The group-relative advantage is computed as
\begin{equation}
\label{eq:grpo-advantage}
A_i^{(k)}
=
\frac{
R_i^{(k)}
-
\operatorname{mean}
\left(
\{R_i^{(j)}\}_{j=1}^{N}
\right)
}{
\operatorname{std}
\left(
\{R_i^{(j)}\}_{j=1}^{N}
\right)
+\epsilon
},
\end{equation}
where $\epsilon$ is a numerical-stability constant. A positive
advantage indicates that the rollout performs better than the group
average, whereas a negative advantage indicates that it performs
worse.

For the $t$-th token of rollout $o_i^{(k)}$, the importance ratio
between the current policy and the rollout policy is
\begin{equation}
\label{eq:grpo-ratio}
r_{i,t}^{(k)}(\theta)
=
\frac{
\pi_\theta
\left(
o_{i,t}^{(k)}
\mid
P_i,o_{i,<t}^{(k)}
\right)
}{
\pi_{\theta_{\mathrm{old}}}
\left(
o_{i,t}^{(k)}
\mid
P_i,o_{i,<t}^{(k)}
\right)
}.
\end{equation}
For prompt $P_i$, the clipped surrogate objective is
\begin{equation}
\label{eq:grpo-objective}
\begin{aligned}
\mathcal{L}_{\mathrm{GRPO},i}(\theta)
={}&
-\frac{1}{N}
\sum_{k=1}^{N}
\frac{1}{|o_i^{(k)}|}
\sum_{t=1}^{|o_i^{(k)}|}
\\
&\min
\Bigl(
r_{i,t}^{(k)}(\theta)A_i^{(k)},
\\
&\qquad
\operatorname{clip}
\bigl(
r_{i,t}^{(k)}(\theta),
1-\epsilon_{\mathrm{clip}},
1+\epsilon_{\mathrm{clip}}
\bigr)
A_i^{(k)}
\Bigr).
\end{aligned}
\end{equation}
The training objective averages
$\mathcal{L}_{\mathrm{GRPO},i}$ over the prompts in the current batch.
The clipping parameter $\epsilon_{\mathrm{clip}}$ constrains the
updated policy from deviating excessively from the rollout policy.
Consistent with our implementation, we optimize this clipped surrogate
objective without an additional reference-policy KL term.

\section{Method}
\label{sec:method}

We start from the task-specific seed set $\mathcal{D}$ and a much
larger external Text-to-SQL pool
\begin{equation}
\label{eq:shared-example-pool}
\mathcal{P} =
\left\{
(q_j^{\mathrm{pool}},
s_j^{\mathrm{pool}},
\mathrm{db}_j^{\mathrm{pool}})
\right\}_{j=1}^{|\mathcal{P}|}.
\end{equation}
In our experiments, $\mathcal{P}$ is the publicly released
SynSQL-2.5M corpus of OmniSQL~\citep{li2025omnisql}.

Figure~\ref{fig:framework} presents the overall framework.
Coverage-Guided Augmentation first identifies SQL structures that are
poorly represented by the seed set and uses them to guide data
synthesis. The resulting examples are combined with $\mathcal{D}$ to
form an augmented training set $\mathcal{D}^{+}$. Failure-Driven
Learning then performs GRPO on $\mathcal{D}^{+}$ and constructs
additional CoT supervision from failures exposed by the evolving
policy. Its two mechanisms serve different purposes: step-level
expansion corrects the failed questions themselves, whereas
epoch-level expansion retrieves new questions that exercise related
SQL structures.

The two stages address different sources of insufficient supervision.
Coverage-Guided Augmentation is performed before policy optimization
and improves the \emph{potential} structural coverage of the training
set. It determines which SQL structures should be introduced when only
a limited number of examples can be added. Failure-Driven Learning is
performed during optimization and reacts to the \emph{actual} behavior
of the current policy. It determines which examples and capabilities
still require direct supervision after the policy has been exposed to
the augmented data. This separation allows the offline construction
stage to remain independent of the evolving student policy, while the
online learning stage adapts to its changing capability boundary.

\subsection{Coverage-Guided Augmentation}
\label{sec:method-aug}

Randomly sampling from a large external pool may repeatedly select SQL
structures that the seed set already covers. Coverage-Guided
Augmentation instead treats the seed skeletons as existing coverage
and selects external skeletons that are far from both the seed set and
previously selected candidates. This favors structures that are
complementary to the seed set and diverse among themselves.

The number of training examples alone is not a reliable measure of
structural coverage. Questions over different databases may
instantiate the same joins, aggregations, or nesting patterns, whereas
a rare but important composition may appear only once. We therefore
measure coverage at the level of schema-independent SQL skeletons
rather than raw examples. Skeleton-level comparison abstracts away
database-specific identifiers while retaining the operators and clause
composition that characterize the reasoning structure of a query.

\paragraph{Candidate Skeletons.}

We extract and deduplicate the skeletons in the seed set and the
external pool:
\begin{equation}
\label{eq:skeleton-sets}
\begin{aligned}
\mathcal{K}_{\mathrm{seed}}
&=
\left\{
\skel(s_i)
:
(q_i,s_i,\mathrm{db}_i)\in\mathcal{D}
\right\},
\\
\mathcal{K}_{\mathrm{pool}}
&=
\left\{
\skel(s_j^{\mathrm{pool}})
:
(q_j^{\mathrm{pool}},
s_j^{\mathrm{pool}},
\mathrm{db}_j^{\mathrm{pool}})
\in\mathcal{P}
\right\}.
\end{aligned}
\end{equation}
We use $\mathcal{K}_{\mathrm{pool}}$ as the candidate collection.
Skeletons that also occur in $\mathcal{K}_{\mathrm{seed}}$ need not
be removed explicitly. Because seed skeletons are treated as existing
centers during selection, exactly covered candidates receive zero
novelty and are naturally deprioritized.

Exact equality, however, captures only one form of redundancy. Two
skeletons may be different strings but still encode closely related
reasoning patterns, such as when they differ only in a projection or
comparison operator. Conversely, a candidate may combine familiar
operators in a genuinely new way. We therefore introduce a continuous
representation that captures graded structural similarity.

\paragraph{Continuous Skeleton Representation.}

We serialize each skeleton and encode it with a fixed encoder
$\emb(\cdot)$. For a skeleton $\kappa$, its normalized
representation is
\begin{equation}
\label{eq:skeleton-embedding}
\mathbf{z}(\kappa) =
\frac{
\emb(\kappa)
}{
\lVert\emb(\kappa)\rVert_2
},
\end{equation}
and the distance between two skeletons is defined as
\begin{equation}
\label{eq:skeleton-distance}
d(\kappa,\kappa') =
1-
\mathbf{z}(\kappa)^\top
\mathbf{z}(\kappa').
\end{equation}
A small distance indicates that the serialized skeletons encode
similar clause and operator compositions, whereas a large distance
indicates stronger structural differences. This provides a graded
notion of structural novelty that distinguishes exact duplicates,
minor variations, and substantially different compositions. The same
representation is later reused for epoch-level retrieval, allowing
both stages to operate in a shared structural space.

\paragraph{Candidate Outlier Filtering.}

A candidate should be far from the seed set because it represents
potentially uncovered structure, but it should not be isolated from
nearly every other pool skeleton. Extreme isolation may indicate a
parsing or representation artifact rather than a meaningful structural
region. We therefore measure whether each candidate is locally
supported by its neighbors.

Let $\mathcal{N}_{k_{\mathrm{nn}}}(\kappa)$ denote the
$k_{\mathrm{nn}}$ nearest pool skeletons to $\kappa$. Its local
isolation score is
\begin{equation}
\label{eq:outlier-score}
\rho(\kappa) =
1-
\frac{1}{k_{\mathrm{nn}}}
\sum_{\kappa'\in
\mathcal{N}_{k_{\mathrm{nn}}}(\kappa)}
\mathbf{z}(\kappa)^\top
\mathbf{z}(\kappa').
\end{equation}
We retain candidates whose isolation score does not exceed a fixed
threshold:
\begin{equation}
\label{eq:filtered-candidate-set}
\mathcal{C} =
\left\{
\kappa\in\mathcal{K}_{\mathrm{pool}}
:
\rho(\kappa)\le\tau_{\mathrm{iso}}
\right\}.
\end{equation}
The concrete values of $k_{\mathrm{nn}}$ and
$\tau_{\mathrm{iso}}$ are reported in the experimental setup.

The filter is intentionally conservative. It does not remove a
candidate merely because it is rare or far from the seed set, since
such candidates may represent precisely the structures that
augmentation should introduce. A candidate is removed only when it is
also weakly supported by its nearest neighbors in the external pool.
The subsequent selection therefore focuses on novel but locally
coherent structural regions rather than isolated points caused by
parsing noise or unstable representations.

\paragraph{Coverage-Guided Skeleton Selection.}

For any candidate $\kappa$ and a set $\mathcal{S}$ of skeletons
already selected for augmentation, we define its distance to the
current coverage as
\begin{equation}
\label{eq:nearest-center-distance}
\Delta(\kappa\mid\mathcal{S}) =
\min_{\kappa'\in
\mathcal{K}_{\mathrm{seed}}\cup\mathcal{S}}
d(\kappa,\kappa').
\end{equation}
A large $\Delta(\kappa\mid\mathcal{S})$ indicates that $\kappa$ is
different from both the seed skeletons and the structures already
selected. Starting with $\mathcal{S}=\emptyset$, we repeatedly add
the candidate with the largest nearest-center distance. Each selected
skeleton becomes a new center, reducing the priority of nearby
candidates in subsequent rounds. This single operation therefore
encourages both complementarity to the seed set and diversity among
the selected structures.

Both parts of the current coverage are necessary. If selection
considered only distance from the seed set, several candidates from
the same uncovered region could all receive high priority and consume
most of the budget. If it considered only diversity among external
candidates, it could select mutually different structures that are
already well represented in the seed set. Treating seed skeletons as
initial centers and every selected skeleton as an additional center
combines the two requirements in one iterative rule.

\paragraph{Construction of Augmented Examples.}

Each selected skeleton serves as a structural template for
synthesizing a task-relevant example on a target training database. For
each $\kappa\in\mathcal{S}$, we make at most $A_{\max}$ attempts
and retain at most one successful example.

Directly copying an example from the external pool would preserve its
original database, schema, and vocabulary, which may be unrelated to
the target task. Instead, we reuse only the selected structural pattern
and instantiate it on databases from the target training set. The
resulting example therefore combines a complementary SQL structure
with schemas, values, and semantics that are relevant to the target
distribution. The construction pipeline separates SQL instantiation,
question generation, and semantic verification so that errors
introduced at one stage do not automatically propagate into the final
training set.

\begin{enumerate}[leftmargin=1.4em,topsep=2pt,itemsep=2pt]
\item \textbf{SQL Instantiation.}
We sample a target database and provide a strong LLM with
$\kappa$, the database schema, column descriptions, sampled
values, and demonstrations from the same database. The generated
SQL $\tilde{s}$ is retained only if it can be parsed, executes
successfully, returns a non-empty result, and preserves the
selected structure:
\[
\skel(\tilde{s})=\kappa.
\]
These checks remove malformed, schema-incompatible, structurally
inconsistent, and semantically vacuous instantiations.

\item \textbf{NL Question Generation.}
Given the accepted SQL and database context, the strong LLM
generates a natural-language question and supporting evidence that
express the intended query without exposing the SQL itself.
Generating the question after SQL instantiation anchors its
semantics to a concrete executable query, while the supporting
evidence allows database-specific terminology and value
interpretation to be stated explicitly when needed.

\item \textbf{Semantic Consistency Verification.}
The strong LLM verifies whether the generated question faithfully
describes the SQL semantics, including the referenced entities,
filtering conditions, aggregation, grouping, ordering, and output
fields. Only examples that pass this verification are retained.
This step filters fluent but semantically inconsistent questions
that omit conditions, reverse ordering requirements, or describe a
different operation from the instantiated SQL.
\end{enumerate}

Let $\mathcal{D}_{\mathrm{aug}}$ denote the set of accepted examples.
Because each selected skeleton contributes at most one example,
\begin{equation}
\label{eq:augmentation-size}
|\mathcal{D}_{\mathrm{aug}}|
\le
K_{\mathrm{sel}}.
\end{equation}
The final training set for policy optimization is
\begin{equation}
\label{eq:augmented-set}
\mathcal{D}^{+} =
\mathcal{D}
\cup
\mathcal{D}_{\mathrm{aug}}.
\end{equation}

The augmented examples are not used in a separate pretraining or SFT
stage. They enter the same policy-optimization pool as the original
seed examples, allowing GRPO to learn from both previously covered and
newly introduced structures under a unified objective.
Coverage-Guided Augmentation therefore changes the structural support
of the training distribution without changing the subsequent
reinforcement-learning procedure.

\subsection{Failure-Driven Learning}
\label{sec:method-rl}

Coverage-Guided Augmentation improves the structural coverage of the
training set, but exposure to an example does not guarantee that the
policy can learn it effectively through GRPO. For some prompts, none
of the $N$ sampled outputs provides a response that is both
well-formatted and execution-correct. Such a rollout group contains no
complete positive behavior for the policy to imitate.

Failure-Driven Learning treats these \emph{Solve-None} examples as
observations of unresolved capabilities. Step-level expansion directly
corrects the failed examples once enough of them have accumulated.
Epoch-level expansion instead aggregates failures observed throughout
an epoch and retrieves additional examples with related SQL
structures. Both mechanisms first identify a supervision batch,
construct and verify CoT traces for that batch, and then apply SFT.

The two temporal scales are complementary. Step-level expansion reacts
quickly to individual failures and supplies the missing positive
trajectory near the training state in which the failure was observed.
Epoch-level expansion operates more slowly: by aggregating failures
across many steps, it identifies recurring capability gaps and
constructs broader practice around the corresponding SQL structures.
The former provides immediate correction, whereas the latter promotes
broader consolidation and transfer.

\subsubsection{Execution Reward and Solve-None Detection}
\label{sec:solve-none}

For each training example, the policy samples
$\{o_i^{(k)}\}_{k=1}^{N}$. We separately evaluate response format and
execution correctness:
\begin{equation}
\label{eq:format-execution-indicators}
\begin{aligned}
F(o)
&=
\mathbf{1}
\left[
o
\text{ follows the required format}
\right],
\\
E(o,s,\mathrm{db})
&=
\mathbf{1}
\left[
\operatorname{extract}(o)
\equiv_{\mathrm{db}}
s
\right].
\end{aligned}
\end{equation}
We further define the complete-correctness indicator
\begin{equation}
\label{eq:complete-correctness}
C(o,s,\mathrm{db}) =
F(o)\,E(o,s,\mathrm{db}),
\end{equation}
which equals one only when the response is both format-valid and
execution-correct. The reward is
\begin{equation}
\label{eq:reward}
R(o,s,\mathrm{db}) =
F(o)
+
\alpha_R C(o,s,\mathrm{db}),
\end{equation}
where $\alpha_R>1$ gives complete execution correctness a larger
contribution than format validity alone.

Let $\mathcal{I}_t$ denote the examples processed at training step
$t$. We define the Solve-None set as
\begin{equation}
\label{eq:solve-none-set}
\mathcal{I}_t^{\mathrm{SN}} =
\left\{
i\in\mathcal{I}_t
:
\sum_{k=1}^{N}
C(o_i^{(k)},s_i,\mathrm{db}_i)
=
0
\right\}.
\end{equation}
In other words, an example is Solve-None if none of its $N$ rollouts
is simultaneously format-valid and execution-correct. This status
depends on the current policy and rollout budget rather than being a
fixed property of the example.

If at least one rollout is fully correct, GRPO observes a complete
positive behavior and can increase its relative probability. If all
rollouts are correct, the example is already solved under the current
sampling budget. In a Solve-None group, however, no rollout provides
the complete target response required for direct imitation. Format
rewards may still distinguish partially valid outputs, but they cannot
demonstrate how to produce the correct SQL. Failure-Driven Learning is
therefore activated for this third case.

\subsubsection{Step-Level Expansion}
\label{sec:step-correction}

Updating on every newly observed failure would produce small and
irregular SFT updates. We therefore store Solve-None examples in a
correction buffer $\mathcal{Q}_{\mathrm{corr}}$ and trigger SFT only
when the buffer contains at least $B_{\mathrm{corr}}$ records.

After the GRPO update at step $t$, if
$|\mathcal{Q}_{\mathrm{corr}}|\ge B_{\mathrm{corr}}$, we remove one
correction batch
\begin{equation}
\label{eq:correction-batch}
\mathcal{J}_t =
\operatorname{Pop}
\left(
\mathcal{Q}_{\mathrm{corr}},
B_{\mathrm{corr}}
\right).
\end{equation}
At most one correction batch is consumed after each GRPO step; any
remaining records stay in the buffer. This ordering preserves GRPO as
the primary update for every mini-batch and prevents a temporary burst
of failures from causing an unbounded sequence of SFT updates at a
single training step. Persistent failures remain in the buffer and can
therefore receive supervision in later steps.

For every $i\in\mathcal{J}_t$, we construct and verify a CoT trace
using Equation~\eqref{eq:cot-trace-construction}. These traces form
\begin{equation}
\label{eq:step-correction-set}
\mathcal{H}_t =
\left\{
(P_i,\tau_i)
:
i\in\mathcal{J}_t
\right\}.
\end{equation}
The policy is updated on $\mathcal{H}_t$ using
$\mathcal{L}_{\mathrm{SFT}}$. Step-level expansion therefore supplies
the complete target behavior missing from the Solve-None rollouts
while retaining GRPO as the primary optimization objective. Because
traces are constructed only after a complete correction batch has
formed, teacher generation and SFT are also performed in regular
batches rather than as one-off operations for individual failures.

\subsubsection{Epoch-Level Expansion}
\label{sec:epoch-practice}

Step-level expansion corrects individual failed questions, but
repeated correction of the same instances may not provide sufficient
practice for the underlying SQL capability. Epoch-level expansion
broadens the supervision in five steps: it aggregates and deduplicates
the failures from an epoch, retrieves question-bank examples with
similar SQL skeletons, gives additional priority to examples matching
diagnosed structural errors, selects one globally budgeted practice
set, and constructs verified CoT traces for the selected examples.

The key distinction is that epoch-level expansion does not simply
replay the same failed questions. It uses each failure as a query into
a larger question bank and retrieves other examples that exercise
related SQL structures. The policy therefore receives multiple
realizations of the same underlying capability across different
questions and databases, encouraging transferable learning rather than
memorization of the original failed instance.

\paragraph{Failure Aggregation.}

For each failed example $i$, let
$u_i=(\mathrm{split}_i,\mathrm{index}_i)$ denote its unique dataset
key. We first aggregate the Solve-None records observed across all
training steps in epoch $e$ into a multiset:
\begin{equation}
\label{eq:raw-epoch-failures}
\widetilde{\mathcal{F}}_e =
\biguplus_{t\in e}
\left\{
i
:
i\in\mathcal{I}_t^{\mathrm{SN}}
\right\}.
\end{equation}
We then deduplicate the records by dataset key:
\begin{equation}
\label{eq:epoch-failures}
\mathcal{F}_e =
\operatorname{UniqueByKey}
\left(
\widetilde{\mathcal{F}}_e;
u_i
\right).
\end{equation}
Here, $\biguplus$ preserves repeated occurrences across training
steps, whereas
$\operatorname{UniqueByKey}(\cdot;u_i)$ retains one record for each
dataset example. This prevents examples that are encountered or fail
multiple times within the same epoch from disproportionately
influencing the subsequent retrieval process. Every distinct failure
therefore contributes one retrieval query. For error-aware processing,
we additionally retain its prompt, gold SQL, and one observed failed
response.

\paragraph{Question Bank and Skeleton Retrieval.}

Epoch-level practice is drawn from an indexed question bank
\begin{equation}
\label{eq:practice-bank}
\mathcal{P}_{\mathrm{prac}} =
\left\{
(q_j^{\mathrm{prac}},
s_j^{\mathrm{prac}},
\mathrm{db}_j^{\mathrm{prac}})
\right\}_{j=1}^{|\mathcal{P}_{\mathrm{prac}}|}.
\end{equation}
Each record contains a question--SQL--database triple and the skeleton
of its gold SQL. Its skeleton collection is
\begin{equation}
\label{eq:practice-skeleton-set}
\mathcal{K}_{\mathrm{prac}} =
\left\{
\skel(s_j^{\mathrm{prac}})
:
1\le j\le|\mathcal{P}_{\mathrm{prac}}|
\right\}.
\end{equation}
Let $\mathcal{U}_{<e}$ denote the indices of question-bank examples
used before epoch $e$.

For each failure $i\in\mathcal{F}_e$, we retrieve its $L$ nearest
bank skeletons:
\begin{equation}
\label{eq:epoch-template-retrieval}
\left(
\nu_{i,1},\ldots,\nu_{i,L}
\right)
=
\operatorname{TopL}_{\nu\in\mathcal{K}_{\mathrm{prac}}}
\mathbf{z}(\skel(s_i))^\top
\mathbf{z}(\nu).
\end{equation}
Candidates retrieved at rank $r$ are aggregated across all failures:
\begin{equation}
\label{eq:rank-candidate-pools}
\mathcal{C}_{e,r} =
\left\{
j
:
\begin{array}{l}
1\le j\le|\mathcal{P}_{\mathrm{prac}}|,\;
j\notin\mathcal{U}_{<e},\\
\exists i\in\mathcal{F}_e,\;
\skel(s_j^{\mathrm{prac}})=\nu_{i,r}
\end{array}
\right\}.
\end{equation}
Failures do not receive separate practice quotas. Instead, they jointly
define rank-stratified candidate pools for one global epoch-level
budget.

We retrieve using the gold SQL skeleton of the failed example rather
than the model-generated SQL. The gold skeleton provides a stable
description of the capability required by the example, whereas the
failed output may be malformed or structurally incomplete. A retrieved
skeleton may correspond to multiple question-bank examples
instantiated on different databases. Aggregating these examples across
failures creates a shared practice pool and allows the available budget
to be spent where suitable examples exist, rather than reserving an
identical quota for every failure.

Retrieval rank is preserved because closer skeletons are generally
more likely to exercise the same composition of clauses and operators.
Lower-ranked pools remain useful for introducing controlled variation
around the failed structure. The global allocation described below
therefore favors higher-ranked pools without restricting practice to
exact or near-exact matches.

\paragraph{Error-Aware Reranking.}

Skeleton similarity provides broad structural practice, while the
failed response may reveal a more specific capability gap. A strong
LLM therefore diagnoses each failure using the structural categories
from NL2SQL-BUGs~\citep{liu2025nl2sqlbugs}. If a structural error is
identified, the failure enters the error-aware branch; otherwise, it
remains covered by standard skeleton retrieval.

For example, if a failed response omits a required subquery, the
error-aware branch prioritizes retrieved examples that exercise the
corresponding subquery structure rather than treating all structurally
similar examples equally. Failures are grouped by diagnosed error
type, and a strong LLM filters and reranks the retrieved candidates for
each group. The total error-aware allocation is capped by
\begin{equation}
\label{eq:error-aware-budget}
K_{\mathrm{ea}} =
\left\lfloor
\gamma K_{\mathrm{ep}}
\right\rfloor,
\end{equation}
where $K_{\mathrm{ep}}$ is the target number of epoch-level practice
examples. We denote the selected error-aware examples by
$\mathcal{B}_e^{\mathrm{ea}}$.

The error-aware branch refines rather than replaces structural
retrieval. Skeleton similarity identifies a broad neighborhood of
relevant examples, while error diagnosis determines which parts of
that neighborhood most directly address the observed mistake. If the
diagnosis is unavailable or uncertain, the failure still contributes
candidates through the standard retrieval branch, preventing the
practice mechanism from depending entirely on a successful error
label.

Grouping failures by error type also allows the practice selection to
reflect the observed distribution of capability gaps. Error patterns
shared by multiple distinct failures can receive greater attention,
while the cap $K_{\mathrm{ea}}$ prevents the error-aware branch from
consuming the entire epoch-level budget.

\paragraph{Global Retrieval Budget.}

The remaining
$(K_{\mathrm{ep}}-|\mathcal{B}_e^{\mathrm{ea}}|)$ slots are allocated
to the rank-stratified pools
$\{\mathcal{C}_{e,r}\}_{r=1}^{L}$ using normalized rank weights
$\boldsymbol{\omega}=(\omega_1,\ldots,\omega_L)$. Any slots left
unfilled because a rank-specific pool contains too few unused
candidates are supplied from the union of the remaining retrieved
examples. Let $\mathcal{B}_e^{\mathrm{ret}}$ denote the selected
standard-retrieval subset. The complete epoch-level practice batch is
\begin{equation}
\label{eq:epoch-practice-batch}
\mathcal{B}_e =
\mathcal{B}_e^{\mathrm{ea}}
\cup
\mathcal{B}_e^{\mathrm{ret}},
\qquad
|\mathcal{B}_e|
\le
K_{\mathrm{ep}}.
\end{equation}
All examples in $\mathcal{B}_e$ originate from skeleton retrieval.
The realized batch may be smaller than $K_{\mathrm{ep}}$ if the
retrieved candidate pools are insufficient. The concrete retrieval
depth, total budget, error-aware ratio, and rank weights are reported
in the experimental setup.

A single global budget makes the amount of epoch-level supervision
predictable even when the number of failures varies across epochs.
The rank weights prioritize examples that are structurally closest to
the observed failures, while the final fill step avoids wasting budget
when a particular rank contains too few candidates. Excluding
previously selected examples further increases practice diversity
across epochs and prevents the auxiliary SFT stage from repeatedly
replaying the same question-bank records.

After selecting $\mathcal{B}_e$, we construct and verify one CoT
trace for each selected example. Let
$P_j^{\mathrm{prac}}$ denote the serialized input constructed from
$(q_j^{\mathrm{prac}},\mathrm{db}_j^{\mathrm{prac}})$. The resulting
epoch-level supervision set is
\begin{equation}
\label{eq:epoch-trajectory-set}
\mathcal{T}_e =
\left\{
(P_j^{\mathrm{prac}},\tau_j)
:
j\in\mathcal{B}_e
\right\}.
\end{equation}
We update the policy on $\mathcal{T}_e$ using
$\mathcal{L}_{\mathrm{SFT}}$, add the selected indices to the set of
previously used bank examples, and use the updated policy in the next
epoch.

Separating example selection from CoT construction clarifies the role
of each component. Retrieval and reranking determine \emph{what} the
policy should practice, whereas the strong model supplies \emph{how}
to solve the selected examples in the required response format.
Constructing traces collectively for the selected batch also applies
the same generation and execution-verification procedure to all
epoch-level supervision.

\paragraph{Overall Training Procedure.}

Algorithm~\ref{alg:failure-driven-learning} summarizes how GRPO
interacts with the two expansion mechanisms. GRPO remains the primary
objective. Step-level expansion periodically corrects buffered
Solve-None examples, whereas epoch-level expansion retrieves broader
practice for the structural capabilities exposed by an epoch of
failures.

\begin{algorithm}[t]
\small
\caption{GRPO with Failure-Driven Learning}
\label{alg:failure-driven-learning}
\begin{algorithmic}[1]
\Require Augmented set $\mathcal{D}^{+}$;
policy $\pi_\theta$;
question bank $\mathcal{P}_{\mathrm{prac}}$;
rollout number $N$;
correction batch size $B_{\mathrm{corr}}$;
epoch practice budget $K_{\mathrm{ep}}$;
number of epochs $E$
\Ensure Updated policy $\pi_\theta$

\State $\mathcal{Q}_{\mathrm{corr}}\gets\emptyset$
\State $\mathcal{U}_{<1}\gets\emptyset$

\For{$(e=1,\ldots,E)$}
\State $\widetilde{\mathcal{F}}_e\gets\emptyset$

\For{each training step $t$ in epoch $e$}
\State Sample rollouts, compute rewards, and identify
$\mathcal{I}_t^{\mathrm{SN}}$
\State Add Solve-None records to
$\mathcal{Q}_{\mathrm{corr}}$ and
$\widetilde{\mathcal{F}}_e$
\State Update $\pi_\theta$ using
$\mathcal{L}_{\mathrm{GRPO}}$

\If{$|\mathcal{Q}_{\mathrm{corr}}|\ge B_{\mathrm{corr}}$}
\State Pop one correction batch and construct
$\mathcal{H}_t$
\State Update $\pi_\theta$ on $\mathcal{H}_t$ using
$\mathcal{L}_{\mathrm{SFT}}$
\EndIf
\EndFor

\State
$
\mathcal{F}_e
\gets
\operatorname{UniqueByKey}
(\widetilde{\mathcal{F}}_e;u_i)
$

\If{$\mathcal{F}_e\neq\emptyset$ and $e<E$}
\State Retrieve and rerank related question-bank examples
\State Select the globally budgeted practice batch
$\mathcal{B}_e$
\State Construct verified CoT traces to form
$\mathcal{T}_e$
\State Update $\pi_\theta$ on $\mathcal{T}_e$ using
$\mathcal{L}_{\mathrm{SFT}}$
\State
$
\mathcal{U}_{<e+1}
\gets
\mathcal{U}_{<e}
\cup
\mathcal{B}_e
$
\Else
\State
$
\mathcal{U}_{<e+1}
\gets
\mathcal{U}_{<e}
$
\EndIf

\EndFor

\State \Return $\pi_\theta$
\end{algorithmic}
\end{algorithm}

\section{Experiments}
\label{sec:experiment}

\begin{table*}[t]
  \centering
  \small
\caption{Performance comparison of different models across Text-to-SQL
benchmarks. Data Eff.\ denotes data efficiency, and Sci.\ Bench.\
denotes Science Benchmark.}
  \label{tab:main}
  \resizebox{\textwidth}{!}{
\begin{tabular}{@{}lcc|cc|cccc@{}}
  \toprule
  \multirow{2}{*}{\textbf{Model}}
  & \multirow{2}{*}{\textbf{Size}}
  & \multirow{2}{*}{\textbf{Train Size}}
  & \multicolumn{2}{c|}{\textbf{Primary}}
  & \multicolumn{4}{c}{\textbf{Generalization}} \\
  \cmidrule(lr){4-5}
  \cmidrule(l){6-9}
  & & &
  \textbf{BIRD Dev}
  & \textbf{Data Eff.}
  & \textbf{Spider Test}
  & \textbf{Spider Dev}
  & \textbf{EHRSQL}
  & \textbf{Sci.\ Bench.} \\
  \midrule
    \multicolumn{9}{c}{\textit{Closed-source LLMs}} \\
    \midrule
    GPT-4o & -- & -- & 61.9 & -- & 83.2 & 70.9 & 44.9 & 55.6 \\
    GPT-4o-mini & -- & -- & 58.8 & -- & 82.4 & 70.4 & 37.9 & 51.8 \\
    GPT-4-Turbo & -- & -- & 62.0 & -- & 83.4 & 72.4 & 43.1 & 59.2 \\
    \midrule
    \multicolumn{9}{c}{\textit{Open-source LLMs}} \\
    \midrule
    Qwen2.5-Coder-7B-Instruct & 7B & -- & 50.9 & -- & 82.5 & 73.4 & 24.3 & 45.2 \\
    Qwen2.5-7B-Instruct & 7B & -- & 46.9 & -- & 76.8 & 65.4 & 20.9 & 38.5 \\
    DeepSeek-Coder-7B-Instruct & 7B & -- & 43.1 & -- & 70.5 & 63.2 & 28.6 & 40.8 \\
    Llama-3.1-8B-Instruct & 8B & -- & 42.0 & -- & 72.2 & 61.8 & 24.6 & 43.1 \\
    OpenCoder-8B-Instruct & 8B & -- & 37.5 & -- & 68.3 & 59.5 & 21.9 & 39.8 \\
    Granite-3.1-8B-Instruct & 8B & -- & 36.0 & -- & 69.8 & 58.3 & 19.6 & 36.8 \\
    Granite-8B-Code-Instruct & 8B & -- & 27.6 & -- & 64.9 & 58.5 & 16.0 & 29.4 \\
    \midrule
    \multicolumn{9}{c}{\textit{Fine-tuned Text-to-SQL Models}} \\
    \midrule
    CodeS & 7B & -- & 57.2 & -- & 80.3 & 72.0 & -- & -- \\
    Share & 3$\times$8B & 41k & 64.1 & 0.53 & 85.9 & 75.3 & -- & -- \\
    OmniSQL & 7B & 2.5M & 63.9 & 0.01 & 87.9 & 81.2 & 34.9 & 50.2 \\
    Reasoning-SQL & 7B & 9.4k & 64.0 & 0.56 & 78.7 & -- & -- & -- \\
    \midrule
    \ours\ (Qwen2.5-Coder-7B-Instruct) & 7B & 12.6k & 64.9 & 1.11 & 85.3 & 75.2 & 36.2 & 53.9 \\
    \ours\ (Llama-3.1-8B-Instruct) & 8B & 12.6k & 61.8 & 1.57 & 83.7 & 72.5 & 34.2 & 49.8 \\
    \bottomrule
  \end{tabular}}
\end{table*}

\subsection{Experimental Setup}
\label{sec:setup}

\paragraph{Benchmarks and metrics.}
We evaluate on multiple benchmarks:
\begin{itemize}[leftmargin=*,topsep=2pt,itemsep=1pt]
  \item \textbf{BIRD}~\citep{li2023bird} includes 12{,}751 NL question--SQL pairs across 95 databases spanning 37 domains, focusing on large-scale noisy data and external knowledge reasoning. We train on its training set and evaluate on the 1{,}534-example development set. BIRD test is hidden and therefore not used.
  \item \textbf{Spider}~\citep{yu2018spider} contains 10{,}181 NL questions and 5{,}693 unique SQL queries, involving 200 databases across multiple tables covering 138 domains. We evaluate on both its development set (1{,}034 examples) and test set (2{,}147 examples).
  \item \textbf{EHRSQL}~\citep{lee2022ehrsql} includes approximately 24{,}000 NL question--SQL pairs linked to 2 open-source electronic health record databases. We evaluate on its 1{,}008-example subset, which introduces challenges such as complex, time-sensitive questions and unanswerable questions.
  \item \textbf{Science Benchmark}~\citep{zhang2023sciencebenchmark} covers research-policy, astrophysics, and cancer-research databases (299 examples).
\end{itemize}
Since our data augmentation is constructed on BIRD databases, Spider, EHRSQL, and Science Benchmark serve as domain generalization benchmarks to assess whether the trained model transfers to unseen database schemas and domains.

We report the following metrics. \textbf{Execution Accuracy (EX)}: a prediction is correct when its execution result matches that of the gold SQL query. Following prior work, we additionally report \textbf{Test-Suite Accuracy (TS)}, an extension of EX that evaluates predictions against a suite of automatically generated test databases, thereby reducing false positives from a single database instance. Since test suites are only available for Spider dev, we report TS on Spider dev and EX on all other benchmarks. \textbf{Data Efficiency}: the BIRD Dev improvement over the base model normalized by the training data scale, defined as $\text{Data Eff.} = \Delta\text{EX} \;/\; (|\mathcal{D}| / 1000)$, where $\Delta\text{EX}$ is the absolute EX gain over the backbone and $|\mathcal{D}|$ is the number of unique training examples.

\paragraph{Baselines.}
We compare against three groups of models. The first group consists of \emph{proprietary models}: GPT-4o~\citep{openai2024gpt4o}, GPT-4o-mini~\citep{openai2024gpt4o}, and GPT-4-Turbo~\citep{openai2023gpt4}. The second group comprises \emph{open-source instruction-tuned models} that are not specialized for Text-to-SQL: Qwen2.5-Coder-7B-Instruct~\citep{hui2024qwen25coder}, Qwen2.5-7B-Instruct~\citep{yang2024qwen2}, DeepSeek-Coder-7B-Instruct~\citep{guo2024deepseek}, Llama-3.1-8B-Instruct~\citep{dubey2024llama3}, OpenCoder-8B-Instruct~\citep{huang2025opencoder}, Granite-3.1-8B-Instruct, and Granite-8B-Code-Instruct~\citep{mishra2024granite}. The third group consists of \emph{Text-to-SQL fine-tuned systems}: CodeS~\citep{li2024codes}, Share~\citep{zhang2024share}, OmniSQL~\citep{li2025omnisql}, and Reasoning-SQL~\citep{pourreza2025reasoningsql}.

\paragraph{Implementation Details.}
We use Qwen3.6-35B-A3B~\citep{yang2025qwen3} as the teacher model for data synthesis, error diagnosis, candidate reranking, and CoT trajectory generation, and Qwen3-Embedding-0.6B~\citep{zhang2025qwen3} for SQL skeleton encoding. We post-train Qwen2.5-Coder-7B-Instruct~\citep{hui2024qwen25coder} and Meta-Llama-3.1-8B-Instruct~\citep{dubey2024llama3} using VeRL~\citep{sheng2024hybridflow} on eight NVIDIA H20 GPUs, with vLLM 0.8.3 for rollout generation.

For Coverage-Guided Augmentation, we set \(k_{\mathrm{nn}}=10\), \(\tau_{\mathrm{iso}}=0.1\), \(K_{\mathrm{sel}}=3{,}500\), and \(A_{\max}=8\). Starting from 3,000 BIRD training examples, the procedure constructs 3,500 synthetic examples, resulting in 6,500 examples for GRPO. GRPO uses a learning rate of \(1\times10^{-6}\), a batch size of 128 prompts, eight rollouts per prompt, a sampling temperature of 0.8, a maximum response length of 2,048 tokens, \(\alpha_R=3\), and 320 training steps.

For step-level expansion, we set \(B_{\mathrm{corr}}=64\) and use an SFT learning rate of \(1\times10^{-6}\). For epoch-level expansion, we set \(L=5\), \(K_{\mathrm{ep}}=1{,}024\), \(\gamma=0.3\), and \(\boldsymbol{\omega}=(0.35,0.25,0.20,0.10,0.10)\), without random sampling. Each selected batch is trained for two SFT epochs with a learning rate of \(1\times10^{-5}\). Epoch-level expansion cumulatively introduces 6,144 question-bank examples, yielding 12,644 distinct task-specific examples in the full run. We use greedy decoding for all evaluations.

\subsection{Main Results}
\label{sec:main-results}

The evaluation results are shown in Table~\ref{tab:main}.

\paragraph{Improvement over the backbone and proprietary models.}
With Qwen2.5-Coder-7B-Instruct as backbone, \ours raises BIRD dev from 50.9 to 64.9 (+14.0), EHRSQL from 24.3 to 36.2 (+11.9), and Science Benchmark from 45.2 to 53.9 (+8.7), improving both in-domain and cross-domain SQL generation from a general-purpose code model. \ours also exceeds proprietary models on BIRD dev, outperforming GPT-4o (61.9), GPT-4-Turbo (62.0), and GPT-4o-mini (58.8). This indicates that targeted post-training on a modest data budget can match or surpass strong proprietary systems on challenging cross-domain Text-to-SQL.

\paragraph{Data efficiency.}
\ours attains the highest BIRD score among the compared systems (64.9) while using only 12.6k unique training examples, yielding a Data Efficiency of 1.11, the highest among all fine-tuned systems. In comparison, OmniSQL reaches 63.9 with 2.5M examples (0.01), Share reaches 64.1 with 41k examples (0.53), and Reasoning-SQL reaches 64.0 with 9.4k examples (0.56). \ours further surpasses OmniSQL on BIRD (64.9 vs.\ 63.9), EHRSQL (36.2 vs.\ 34.9), and Science Benchmark (53.9 vs.\ 50.2) using roughly $200\times$ less training data, suggesting that coverage-guided augmentation and failure-driven learning extract more value per example than large-scale indiscriminate synthesis.

\paragraph{Generalization across backbones.}
Replacing the backbone with the weaker Llama-3.1-8B-Instruct (BIRD 42.0), the same pipeline yields 61.8 on BIRD (+19.8 over base) and consistent improvements across the generalization benchmarks, with a Data Efficiency of 1.57. The larger relative gain on a weaker starting point indicates that our framework is not tied to a specific backbone and transfers across architectures.

\subsection{Ablation Study}
\label{sec:ablation}

Table~\ref{tab:ablation} evaluates the contribution of each component of \ours\ using Qwen2.5-Coder-7B-Instruct as the backbone.

\paragraph{Optimization and online supervision.}
The reference configurations first demonstrate the importance of the training strategy. SFT on the seed set achieves 55.8 on BIRD dev, while GRPO on the augmented training set reaches 63.2. Applying pure SFT to all available training examples improves the result to 60.5 but remains 4.4 points below the full framework, indicating that simply increasing supervised data is insufficient to match the combination of execution-guided optimization and dynamically constructed supervision.

The two failure-driven mechanisms provide further improvements over GRPO. Adding step-level correction to the RL-only configuration increases BIRD dev from 63.2 to 63.8, and further introducing epoch-level expansion raises it to 64.9. Conversely, removing step-level correction from the full framework reduces the score to 63.4. These comparisons show that step-level correction and epoch-level expansion address complementary aspects of Solve-None failures, although their effects are not necessarily additive because each mechanism changes the policy and hence the failures subsequently observed during training.

\paragraph{Failure-aware practice selection.}
Replacing failure-driven epoch-level retrieval with random selection from the same practice pool reduces BIRD dev from 64.9 to 64.3. The differences are larger on Spider Test and Science Benchmark, where failure-aware selection improves performance by 1.5 and 5.7 points, respectively. This indicates that retrieving practice according to the policy's observed structural weaknesses is more effective than spending the same epoch-level practice budget on randomly selected examples. The particularly large gain on Science Benchmark suggests that failure-aware practice strengthens SQL capabilities that transfer beyond the training databases.

\paragraph{Coverage-guided augmentation.}
Replacing (K)-center skeleton selection with random augmentation from the same external reservoir lowers performance from 64.9 to 64.4 on BIRD dev, from 85.3 to 83.4 on Spider Test, and from 53.9 to 49.8 on Science Benchmark. Although the in-domain improvement is moderate, the larger gains on the two generalization benchmarks indicate that selecting structures complementary to the seed set produces more transferable supervision than randomly expanding the training data.

\begin{table}[t]
  \centering
  \small
  \caption{Results of ablation study (\%). Experiments are conducted on Qwen2.5-Coder-7B-Instruct.}
  \label{tab:ablation}
  \begin{tabular}{@{}lccc@{}}
    \toprule
    \textbf{Configuration} & \textbf{BIRD Dev} & \textbf{Spider Test} & \textbf{Sci.\ Bench.} \\
    \midrule
    \textbf{Full \ours} & \textbf{64.9} & \textbf{85.3} & \textbf{53.9} \\
    \midrule
    SFT (seed data) & 55.8 & 81.5 & 46.2 \\
    SFT (all data) & 60.5 & 82.3 & 48.5 \\
    RL only & 63.2 & 84.7 & 49.5 \\
    w/o Epoch-Level Extra SFT & 63.8 & 85.4 & 51.5 \\
    w/o Step-Level Correction & 63.4 & 83.7 & 47.8 \\
    w/ Random Epoch-Level SFT & 64.3 & 83.8 & 48.2 \\
    w/ Random Augmentation & 64.4 & 83.4 & 49.8 \\
    \bottomrule
  \end{tabular}
\end{table}

\subsection{Augmentation Efficiency and Coverage}
\label{sec:aug-efficiency}

To quantify how effectively each augmentation strategy covers the structural space of the evaluation workload, we measure skeleton $n$-gram coverage on BIRD dev. Let $V$ be the set of unique keyword $n$-grams extracted from BIRD dev skeletons. Given a training set $D$, we define:
\begin{equation}
  \text{Cov}(D) = \frac{|V \cap \text{n-grams}(D)|}{|V|}
\end{equation}
We further define \emph{Recovery} as the fraction of the uncovered structural space that the augmentation fills:
\begin{equation}
  \text{Recovery} = \frac{\Delta\text{Cov}}{1 - \text{Cov(seed)}}
\end{equation}
where $\Delta\text{Cov} = \text{Cov}(\text{seed} \cup \text{aug}) - \text{Cov}(\text{seed})$. Starting from the same 3{,}000 BIRD seed examples, we compare our $K$-Center skeleton-guided augmentation with OmniSQL-style augmentation. For the OmniSQL baseline, we follow its data construction pipeline to synthesize examples on the same BIRD databases, producing 3{,}500 examples equal in size to ours.

\begin{table}[t]
  \centering
  \small
  \caption{Structural coverage efficiency on BIRD dev.}
  \label{tab:aug-efficiency}
  \resizebox{\columnwidth}{!}{%
  \begin{tabular}{@{}ccccc@{}}
    \toprule
    $n$ & Seed Cov. & Method & Coverage after Aug. & Recovery \\
    \midrule
    \multirow{2}{*}{1} & \multirow{2}{*}{79.7\%} & OmniSQL & 94.6\% {\scriptsize($\uparrow$14.9)} & 73.3\% \\
    & & \textbf{Ours} & \textbf{100.0\%} {\scriptsize($\uparrow$20.3)} & \textbf{100.0\%} \\
    \midrule
    \multirow{2}{*}{2} & \multirow{2}{*}{74.5\%} & OmniSQL & 91.0\% {\scriptsize($\uparrow$16.6)} & 64.9\% \\
    & & \textbf{Ours} & \textbf{95.2\%} {\scriptsize($\uparrow$20.7)} & \textbf{81.1\%} \\
    \midrule
    \multirow{2}{*}{3} & \multirow{2}{*}{69.4\%} & OmniSQL & 87.0\% {\scriptsize($\uparrow$17.5)} & 57.3\% \\
    & & \textbf{Ours} & \textbf{91.6\%} {\scriptsize($\uparrow$22.2)} & \textbf{72.5\%} \\
    \bottomrule
  \end{tabular}%
  }
\end{table}

As shown in Table~\ref{tab:aug-efficiency}, our method achieves higher coverage and recovery at all \(n\)-gram levels. At \(n=1\), our augmentation reaches complete coverage of the SQL keywords observed in BIRD dev, recovering all structural unigrams missing from the seed set. More importantly, the advantage persists for higher-order \(n\)-grams: our method improves Recovery over OmniSQL by 16.2 points at \(n=2\) and 15.2 points at \(n=3\). Since these higher-order units capture local combinations of clauses and operators, the results indicate that coverage-guided selection introduces not only previously missing SQL components but also more diverse structural compositions. Moreover, the \(n\)-gram metric is not directly optimized by the embedding-based \(K\)-center procedure, providing complementary evidence that the selected skeletons expand the structural support of the training set under the same augmentation budget.

\begin{figure}[t]
  \centering
  \includegraphics[width=\columnwidth]{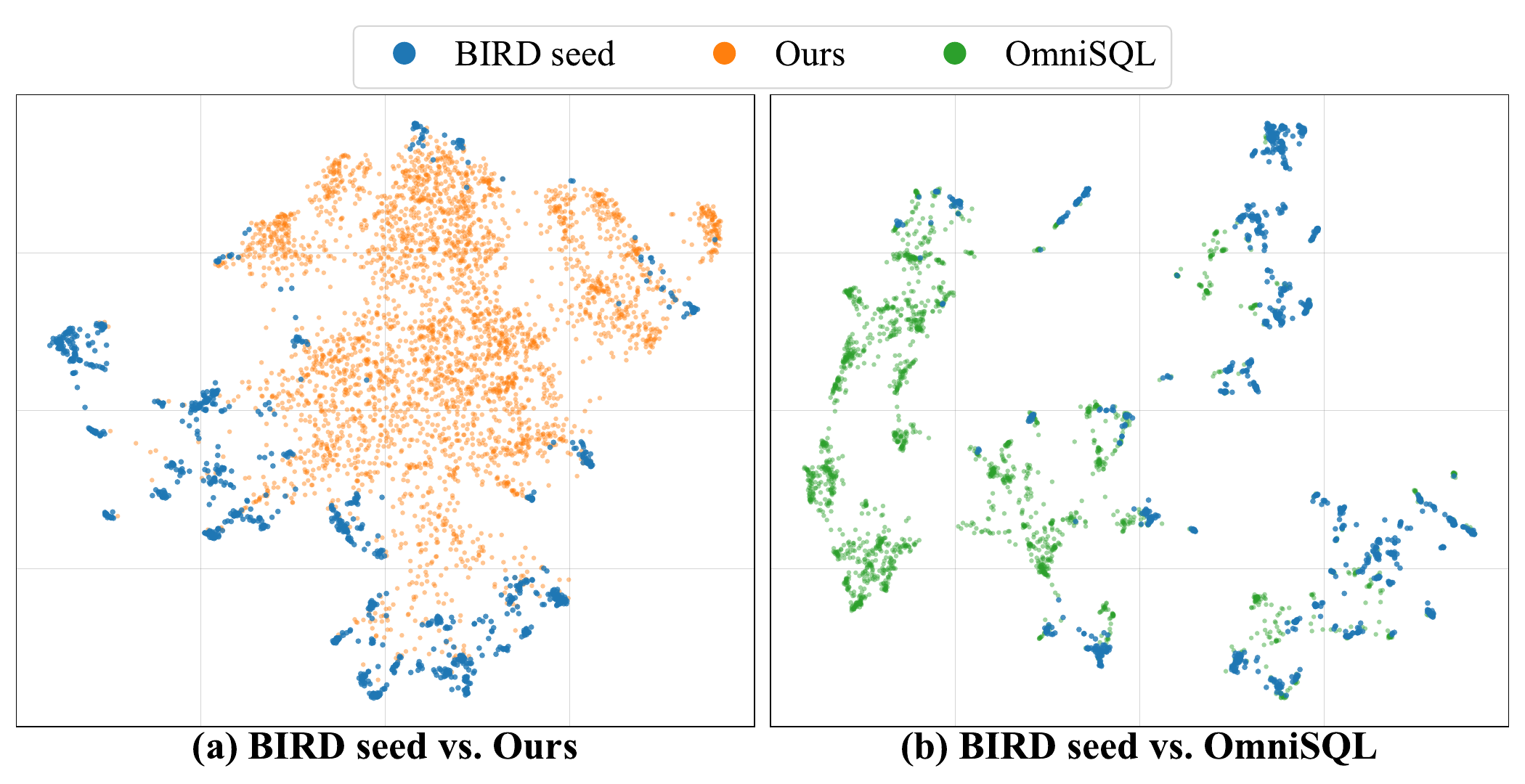}
  \caption{UMAP of SQL skeleton embeddings. (a) BIRD seed vs.\ our augmentation. (b) BIRD seed vs.\ OmniSQL augmentation. Both augmentations produce 3{,}500 examples from the same seed.}
  \label{fig:skeleton-umap}
\end{figure}

Figure~\ref{fig:skeleton-umap} provides a qualitative visualization of the coverage difference. Panel~(a) shows that our augmented examples spread broadly across the skeleton embedding space, filling regions that are sparse under the BIRD seed. In contrast, panel~(b) shows that OmniSQL augmentation tends to cluster in a subset of regions, leaving many areas of the seed distribution uncovered. This pattern is consistent with the quantitative coverage gap reported in Table~\ref{tab:aug-efficiency}.

\subsection{Training Dynamics and Cost}
\label{sec:cost}

\paragraph{Cost.}
Constructing one accepted example requires 7.67 LLM calls, 36.8k input tokens, and 4.1k output tokens on average. The complete training run, including GRPO, step-level correction, and epoch-level Extra SFT, takes 23.47 hours on 8 H20 GPUs. The skeleton-based practice retrieval adds negligible overhead: each epoch-level retrieval over the reservoir of roughly 74k skeletons takes 2--3 seconds on CPU (about 15 seconds in total, under 0.02\% of end-to-end time), and the index is loaded once without consuming GPU memory. Failure-driven learning therefore improves data efficiency without introducing a meaningful computational cost.

\begin{figure*}[t]
  \centering
  \includegraphics[width=0.32\textwidth]{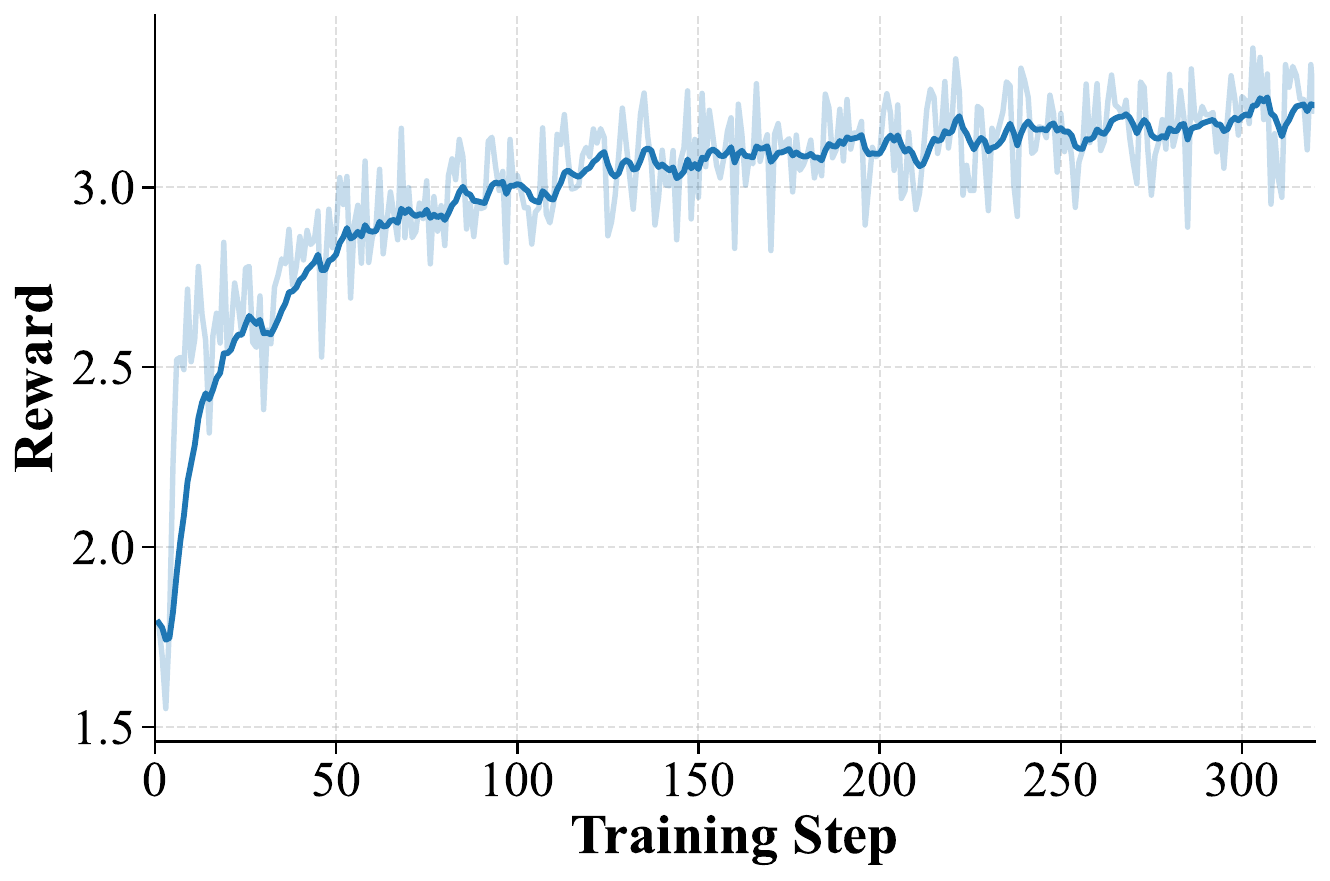}\hfill
  \includegraphics[width=0.32\textwidth]{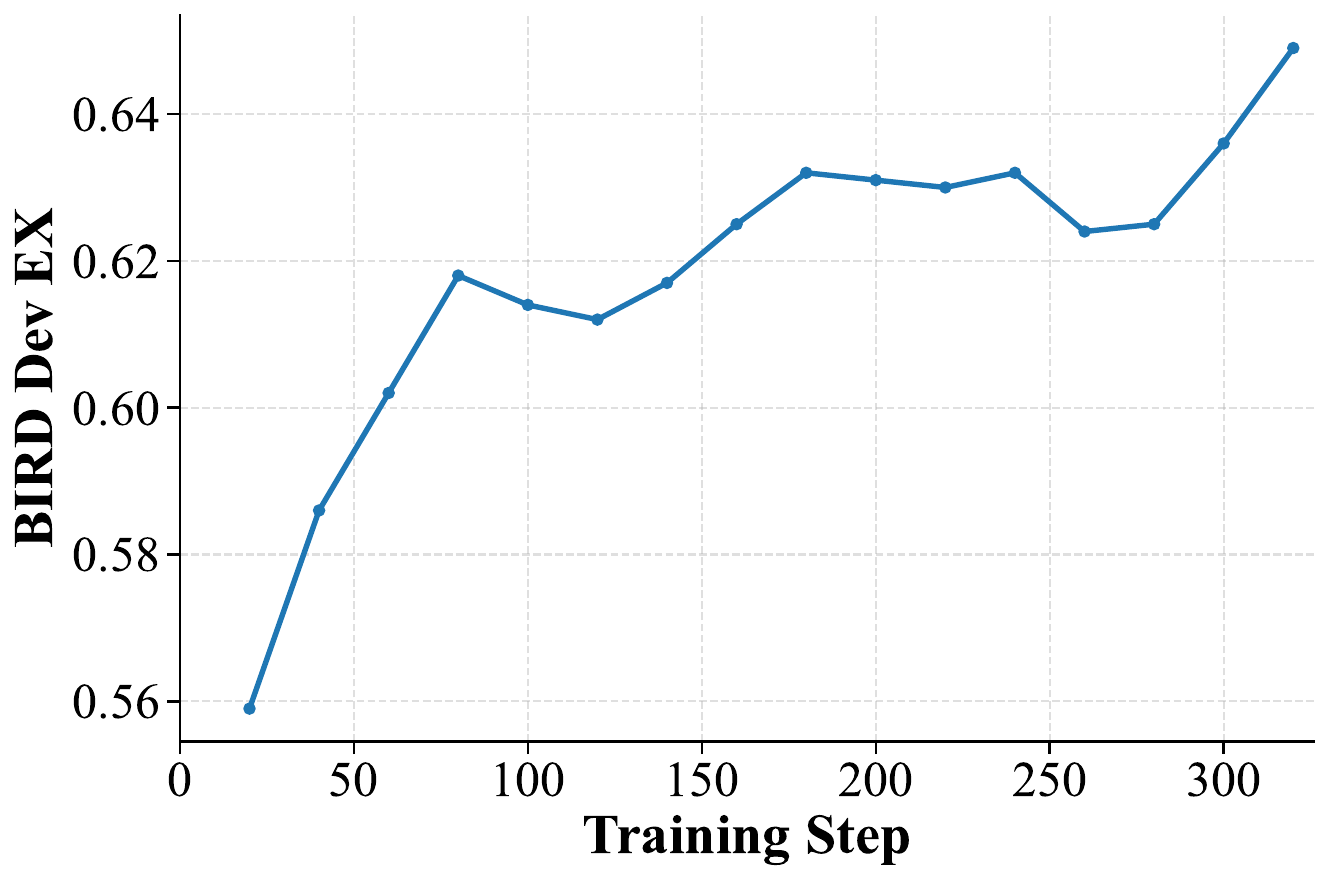}\hfill
  \includegraphics[width=0.32\textwidth]{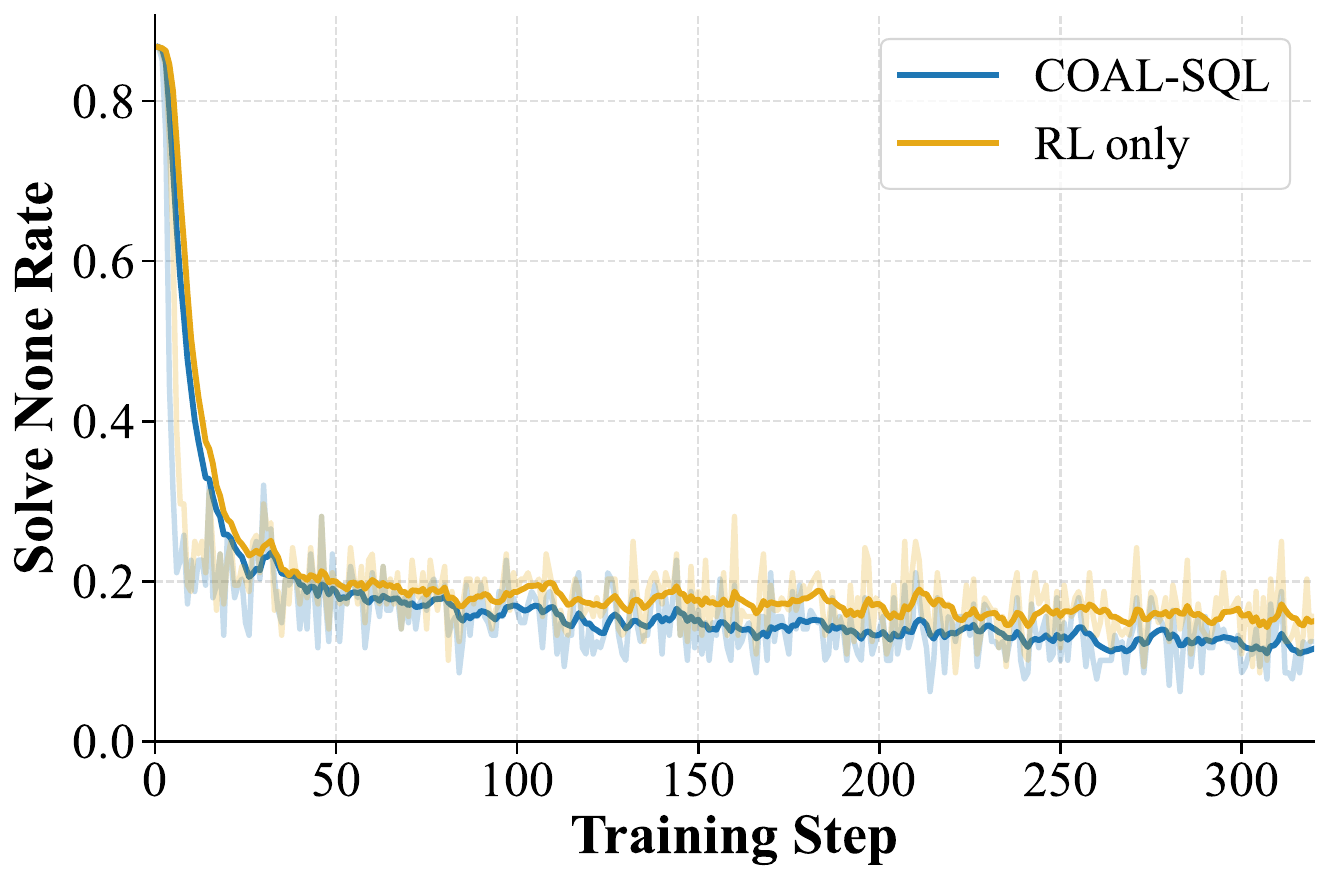}
  \caption{Training dynamics. Left: mean reward. Middle: BIRD dev execution accuracy. Right: \solveNone rate for full \ours and the RL-only baseline.}
  \Description{Three line charts showing training dynamics. Left: mean reward rises from about 1.6 to 3.3. Middle: BIRD dev accuracy for COAL-SQL consistently exceeds RL-only throughout training. Right: solve-none rates for COAL-SQL and RL-only both start near 0.86 and fall, with COAL-SQL generally lower.}
  \label{fig:training-dynamics}
\end{figure*}

\paragraph{Training dynamics.}
Figure~\ref{fig:training-dynamics} tracks three metrics throughout training. The mean reward (left) rises from about 1.6 to approximately 3.2--3.3. The BIRD dev accuracy (middle) shows that full \ours reaches higher EX than RL only throughout training, confirming that the online SFT stages translate into improved held-out performance. The \solveNone rate (right), i.e., the fraction of prompts for which none of the $N=8$ rollouts is execution-correct, falls rapidly for both runs but remains generally lower for full \ours than for the RL-only baseline, consistent with failure-driven correction reducing all-incorrect rollout groups. Because these are single-run traces with substantial step-to-step variation, they are descriptive rather than an isolated causal estimate for any individual component.

\begin{figure}[t]
  \centering
  \includegraphics[width=0.45\textwidth]{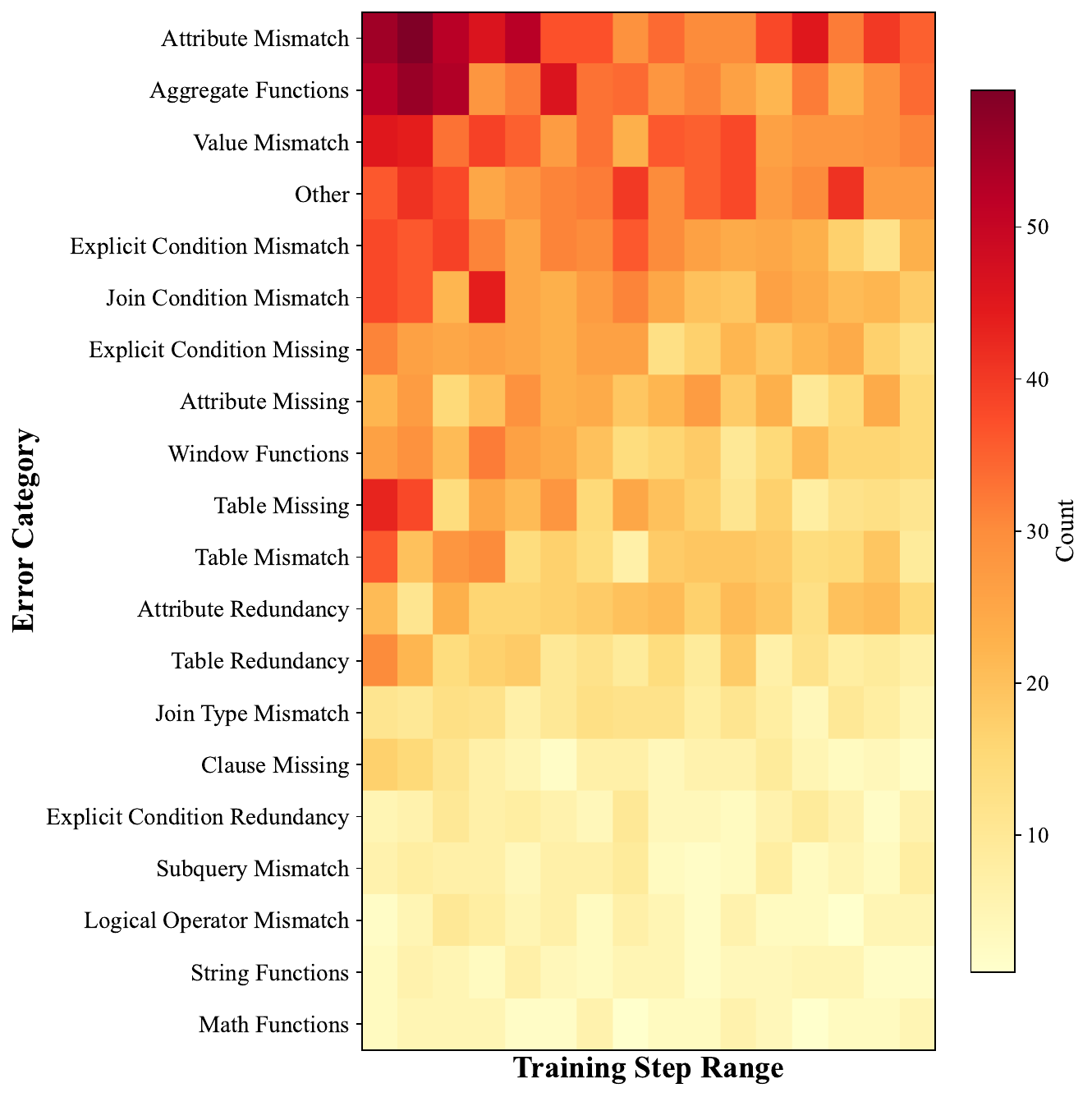}
  \caption{Heatmap of the top-20 error types on incorrect rollouts across training, classified by the NL2SQL-BUGs taxonomy. Darker cells indicate higher error frequency.}
  \Description{A heatmap showing the evolution of different error types over training steps. Structural errors decrease rapidly in the early phase, while schema-grounding errors decrease more gradually throughout training.}
  \label{fig:error-heatmap}
\end{figure}

\paragraph{Error-type evolution during training.}
To further characterize the learning process, Figure~\ref{fig:error-heatmap} tracks the count of each error type across training steps for questions on which the policy produces incorrect rollouts. Error types follow the NL2SQL-BUGs taxonomy~\citep{liu2025nl2sqlbugs}. Structural errors such as incorrect joins, aggregations, and nested queries decrease sharply during the early phase, as execution-reward optimization directly penalizes compositionally incorrect SQL. In contrast, schema-grounding errors involving incorrect tables, columns, or values decline more gradually throughout training, suggesting that these finer-grained mistakes require the targeted corrective signals from failure-driven learning—step-level correction and failure-aware practice expansion—to be effectively addressed.

\subsection{Synthetic-Data Quality and Contamination}
\label{sec:data-quality}

\paragraph{Construction yield.}
We report the acceptance rate at each construction stage. In SQL Instantiation, 29.9\% of generated queries execute successfully, and 78.1\% of the executable ones return a non-empty result; after all instantiation gates (parsing, execution, non-empty result, and skeleton match), the overall acceptance rate is 22.6\%. In Semantic Consistency Verification, 84.7\% of the generated question--SQL pairs are judged to faithfully express the query intent. This last rate reflects an LLM-based semantic judgment rather than human annotation, which we complement with the human evaluation below.

\paragraph{LLM-based data quality assessment.}
To assess the quality of our synthetic data, we randomly sample 200 accepted examples and use DeepSeek-V4-Flash as an LLM judge to score them along five dimensions: \emph{Question-SQL Consistency}, \emph{Result Precision}, \emph{Logical Alignment}, \emph{Query Validity}, and \emph{Linguistic Fluency}. For comparison, we apply the same protocol to 200 randomly sampled BIRD training examples. Following OmniSQL~\citep{li2025omnisql}, the judge rates each criterion as \emph{excellent}, \emph{good}, \emph{average}, or \emph{poor}, and we aggregate the ratings into a weighted score:
\begin{equation}
  \text{Score} = \frac{N_e \times 1.0 + N_g \times 0.75 + N_a \times 0.5 + N_p \times 0.25}{N_e + N_g + N_a + N_p} \times 100,
\end{equation}
where $N_e$, $N_g$, $N_a$, and $N_p$ are the numbers of examples rated as excellent, good, average, and poor, respectively.

\begin{figure}[t]
  \centering
  \includegraphics[width=0.85\columnwidth]{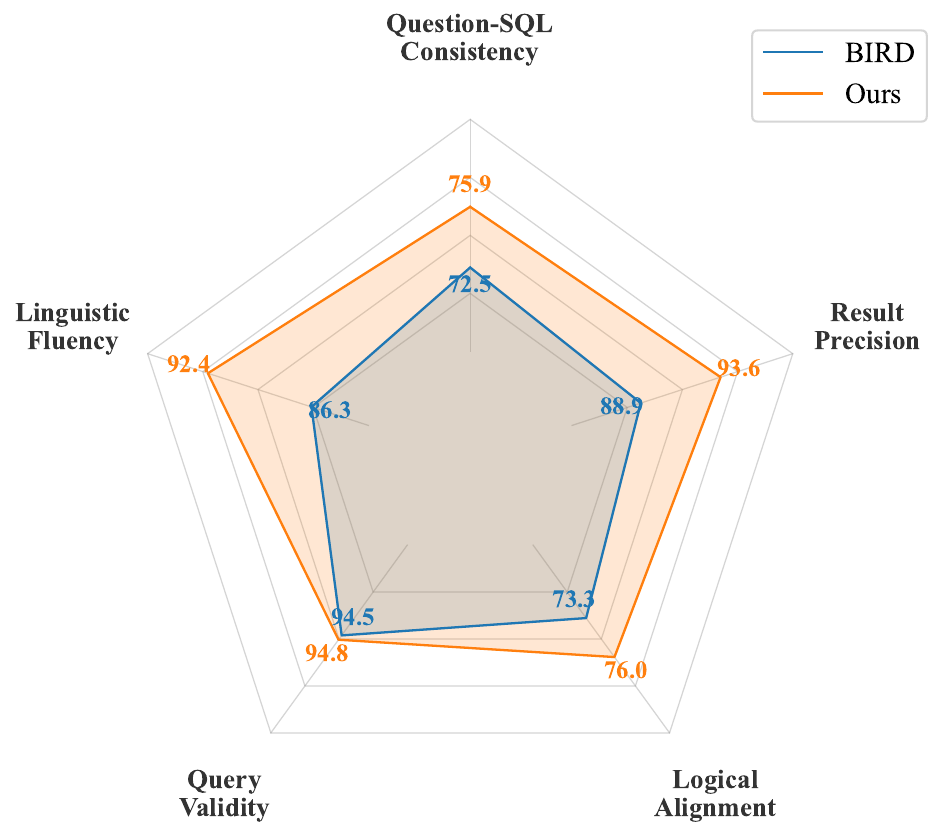}
  \caption{LLM-based data quality assessment (0--100) across five dimensions, comparing our synthetic data with BIRD training data.}
  \label{fig:data-quality-radar}
\end{figure}

Figure~\ref{fig:data-quality-radar} shows that our synthetic data matches or exceeds the human-annotated BIRD data on all five dimensions. The gains are most pronounced on Linguistic Fluency (92.4 vs.\ 86.3) and Result Precision (93.6 vs.\ 88.9), while Query Validity is comparably high for both (94.8 vs.\ 94.5). These results indicate that our automatically constructed data attains a quality on par with human annotation, without introducing systematic degradation.

\paragraph{Data contamination analysis.}
Since our training data is constructed on BIRD databases, we audit the overlap between the 12{,}644 training questions and the 1{,}534 BIRD dev questions. For natural-language questions, we find no exact duplicate and no shared 13-gram; only 2.09\% of dev questions share even an 8-gram with any training question. For SQL, no dev query reaches a token-level containment of 0.5 against any training query, and a MinHash audit finds no pair with Jaccard similarity above 0.71. These results show no substantial train--dev overlap, indicating that the observed gains are not attributable to contamination.

\section{Conclusion}

We proposed \ours, a unified framework that improves structural coverage and adaptive supervision for task-specific Text-to-SQL post-training. It combines Coverage-Guided Augmentation with Failure-Driven Learning, which supplements GRPO through step-level correction and epoch-level practice. Using 12{,}644 distinct examples, \ours achieves 64.9\% execution accuracy on BIRD dev with Qwen2.5-Coder-7B-Instruct and generalizes well across benchmarks. Ablation studies confirm the effectiveness and complementarity of the proposed components.



\bibliographystyle{ACM-Reference-Format}
\bibliography{sample}

@String{Computing = "Computing" }

@ArtifactSoftware{R,
    title = {R: A Language and Environment for Statistical Computing},
    author = {{R Core Team}},
    organization = {R Foundation for Statistical Computing},
    address = {Vienna, Austria},
    year = {2019},
    url = {https://www.R-project.org/},
}

@inproceedings{yu2018spider,
  title     = {{Spider}: A Large-Scale Human-Labeled Dataset for Complex and Cross-Domain Semantic Parsing and Text-to-{SQL} Task},
  author    = {Yu, Tao and Zhang, Rui and Yang, Kai and Yasunaga, Michihiro and Wang, Dongxu and Li, Zifan and Ma, James and Li, Irene and Yao, Qingning and Roman, Shanelle and Zhang, Zilin and Radev, Dragomir},
  booktitle = {Proceedings of the 2018 Conference on Empirical Methods in Natural Language Processing ({EMNLP})},
  year      = {2018}
}

@inproceedings{li2023bird,
  title     = {Can {LLM} Already Serve as A Database Interface? {A} {BI}g Bench for Large-Scale Database Grounded Text-to-{SQLs}},
  author    = {Li, Jinyang and Hui, Binyuan and Qu, Ge and Yang, Jiaxi and Li, Binhua and Li, Bowen and Wang, Bailin and Qin, Bowen and Geng, Ruiyang and Huo, Nan and Zhou, Xuanhe and Ma, Chenhao and Li, Guoliang and Chang, Kevin C.-C. and Huang, Fei and Cheng, Reynold and Li, Yongbin},
  booktitle = {Advances in Neural Information Processing Systems ({NeurIPS})},
  year      = {2023}
}

@article{gao2024dailsql,
  title   = {Text-to-{SQL} Empowered by Large Language Models: {A} Benchmark Evaluation},
  author  = {Gao, Dawei and Wang, Haibin and Li, Yaliang and Sun, Xiuyu and Qian, Yichen and Ding, Bolin and Zhou, Jingren},
  journal = {Proceedings of the {VLDB} Endowment},
  volume  = {17},
  number  = {5},
  pages   = {1132--1145},
  year    = {2024}
}

@article{li2025omnisql,
  title   = {{OmniSQL}: Synthesizing High-Quality Text-to-{SQL} Data at Scale},
  author  = {Li, Haoyang and Wu, Shang and Zhang, Xiaokang and Huang, Xinmei and Zhang, Jing and Jiang, Fuxin and Wang, Shuai and Zhang, Tieying and Chen, Jianjun and Shi, Rui and Chen, Hong and Li, Cuiping},
  journal = {Proceedings of the {VLDB} Endowment},
  volume  = {18},
  number  = {11},
  year    = {2025},
  note    = {arXiv:2503.02240}
}

@inproceedings{liu2025nl2sqlbugs,
  title     = {{NL2SQL-BUGs}: A Benchmark for Detecting Semantic Errors in {NL2SQL} Translation},
  author    = {Liu, Xinyu and Shen, Shuyu and Li, Boyan and Tang, Nan and Luo, Yuyu},
  booktitle = {Proceedings of the 31st {ACM} {SIGKDD} Conference on Knowledge Discovery and Data Mining ({KDD})},
  year      = {2025},
  note      = {arXiv:2503.11984}
}

@inproceedings{pourreza2023dinsql,
  title     = {{DIN-SQL}: Decomposed In-Context Learning of Text-to-{SQL} with Self-Correction},
  author    = {Pourreza, Mohammadreza and Rafiei, Davood},
  booktitle = {Advances in Neural Information Processing Systems ({NeurIPS})},
  year      = {2023}
}

@inproceedings{scholak2021picard,
  title     = {{PICARD}: Parsing Incrementally for Constrained Auto-Regressive Decoding from Language Models},
  author    = {Scholak, Torsten and Schucher, Nathan and Bahdanau, Dzmitry},
  booktitle = {Proceedings of the 2021 Conference on Empirical Methods in Natural Language Processing ({EMNLP})},
  year      = {2021}
}

@inproceedings{zhong2017seq2sql,
  title     = {{Seq2SQL}: Generating Structured Queries from Natural Language using Reinforcement Learning},
  author    = {Zhong, Victor and Xiong, Caiming and Socher, Richard},
  booktitle = {International Conference on Learning Representations ({ICLR})},
  year      = {2018}
}

@inproceedings{li2023resdsql,
  title     = {{RESDSQL}: Decoupling Schema Linking and Skeleton Parsing for Text-to-{SQL}},
  author    = {Li, Haoyang and Zhang, Jing and Li, Cuiping and Chen, Hong},
  booktitle = {Proceedings of the AAAI Conference on Artificial Intelligence ({AAAI})},
  volume    = {37},
  number    = {11},
  pages     = {13067--13075},
  year      = {2023}
}

@article{shao2024deepseekmath,
  title   = {{DeepSeekMath}: Pushing the Limits of Mathematical Reasoning in Open Language Models},
  author  = {Shao, Zhihong and Wang, Peiyi and Zhu, Qihao and Xu, Runxin and Song, Junxiao and Bi, Xiao and Zhang, Haowei and Zhang, Mingchuan and Li, Y. K. and Wu, Y. and Guo, Daya},
  journal = {arXiv preprint arXiv:2402.03300},
  year    = {2024}
}

@article{schulman2017ppo,
  title   = {Proximal Policy Optimization Algorithms},
  author  = {Schulman, John and Wolski, Filip and Dhariwal, Prafulla and Radford, Alec and Klimov, Oleg},
  journal = {arXiv preprint arXiv:1707.06347},
  year    = {2017}
}

@inproceedings{yan2025luffy,
  title     = {Learning to Reason under Off-Policy Guidance},
  author    = {Yan, Jianhao and Li, Yafu and Hu, Zican and Wang, Zhi and Cui, Ganqu and Qu, Xiaoye and Cheng, Yu and Zhang, Yue},
  booktitle = {Advances in Neural Information Processing Systems ({NeurIPS})},
  year      = {2025},
  note      = {arXiv:2504.14945}
}

@article{ma2025relift,
  title   = {Learning What Reinforcement Learning Can't: {I}nterleaved Online Fine-Tuning for Hardest Questions},
  author  = {Ma, Lu and Liang, Hao and Qiang, Meiyi and Tang, Lexiang and Ma, Xiaochen and Wong, Zhen Hao and Niu, Junbo and Shen, Chengyu and He, Runming and Li, Yanhao and Cui, Bin and Zhang, Wentao},
  journal = {arXiv preprint arXiv:2506.07527},
  year    = {2025}
}

@inproceedings{luong2024reft,
  title     = {{ReFT}: Reasoning with Reinforced Fine-Tuning},
  author    = {Luong, Trung Quoc and Zhang, Xinbo and Jie, Zhanming and Sun, Peng and Jin, Xiaoran and Li, Hang},
  booktitle = {Proceedings of the 62nd Annual Meeting of the Association for Computational Linguistics ({ACL})},
  year      = {2024}
}

@article{hui2024qwen25coder,
  title   = {{Qwen2.5-Coder} Technical Report},
  author  = {Hui, Binyuan and Yang, Jian and Cui, Zeyu and Yang, Jiaxi and Liu, Dayiheng and Zhang, Lei and Liu, Tianyu and Zhang, Jiajun and Yu, Bowen and Lu, Keming and Dang, Kai and Fan, Yang and Zhang, Yichang and Yang, An and Men, Rui and Huang, Fei and Zheng, Bo and Miao, Yibo and Quan, Shanghaoran and Feng, Yunlong and Ren, Xingzhang and Ren, Xuancheng and Zhou, Jingren and Lin, Junyang},
  journal = {arXiv preprint arXiv:2409.12186},
  year    = {2024}
}

@inproceedings{wei2022cot,
  title     = {Chain-of-Thought Prompting Elicits Reasoning in Large Language Models},
  author    = {Wei, Jason and Wang, Xuezhi and Schuurmans, Dale and Bosma, Maarten and Ichter, Brian and Xia, Fei and Chi, Ed H. and Le, Quoc V. and Zhou, Denny},
  booktitle = {Advances in Neural Information Processing Systems ({NeurIPS})},
  year      = {2022}
}

@inproceedings{li2024codes,
  title     = {{CodeS}: Towards Building Open-source Language Models for Text-to-SQL},
  author    = {Li, Haoyang and Zhang, Jing and Liu, Hanbing and Fan, Ju and Zhang, Xiaokang and Zhu, Jun and Wei, Renjie and Pan, Hongyan and Li, Cuiping and Chen, Hong},
  booktitle = {Proceedings of the ACM on Management of Data (SIGMOD)},
  volume    = {2},
  number    = {3},
  year      = {2024}
}

@inproceedings{lee2022ehrsql,
  title     = {{EHRSQL}: A Practical Text-to-SQL Benchmark for Electronic Health Records},
  author    = {Lee, Gyubok and Hwang, Hyeonji and Bae, Seongsu and Kwon, Yeonsu and Shin, Woncheol and Yang, Seongjun and Seo, Minjoon and Kim, Jong-Yeup and Choi, Edward},
  booktitle = {Advances in Neural Information Processing Systems (NeurIPS) Datasets and Benchmarks Track},
  volume    = {35},
  year      = {2022}
}

@article{liu2024nl2sqlsurvey,
  title   = {A Survey of Text-to-SQL in the Era of {LLMs}: Where are we, and where are we going?},
  author  = {Liu, Xinyu and Shen, Shuyu and Li, Boyan and Ma, Peixian and Jiang, Runzhi and Zhang, Yuxin and Fan, Ju and Li, Guoliang and Tang, Nan and Luo, Yuyu},
  journal = {IEEE Transactions on Knowledge and Data Engineering (TKDE)},
  year    = {2025}
}

@article{pourreza2025reasoningsql,
  title   = {Reasoning-SQL: Reinforcement Learning with SQL Tailored Partial Rewards for Reasoning-Enhanced Text-to-SQL},
  author  = {Pourreza, Mohammadreza and Talaei, Shayan and Sun, Ruoxi and Wan, Xingchen and Li, Hailong and Mirhoseini, Azalia and Saberi, Amin and Arik, Sercan O.},
  journal = {arXiv preprint arXiv:2503.23157},
  year    = {2025}
}

@inproceedings{zhang2024share,
  title     = {{SHARE}: An {SLM}-based Hierarchical Action CorREction Assistant for Text-to-SQL},
  author    = {Qu, Ge and Li, Jinyang and Qin, Bowen and Li, Xiaolong and Huo, Nan and Ma, Chenhao and Cheng, Reynold},
  booktitle = {Proceedings of the 63rd Annual Meeting of the Association for Computational Linguistics (ACL)},
  year      = {2025}
}

@article{dubey2024llama3,
  title   = {The Llama 3 Herd of Models},
  author  = {Dubey, Abhimanyu and others},
  journal = {arXiv preprint arXiv:2407.21783},
  year    = {2024}
}

@article{openai2024gpt4o,
  title   = {{GPT-4o} System Card},
  author  = {{OpenAI}},
  journal = {arXiv preprint arXiv:2410.21276},
  year    = {2024}
}

@article{openai2023gpt4,
  title   = {{GPT-4} Technical Report},
  author  = {{OpenAI}},
  journal = {arXiv preprint arXiv:2303.08774},
  year    = {2023}
}

@article{yao2025arctic,
  title   = {{Arctic-Text2SQL-R1}: Simple Rewards, Strong Reasoning in Text-to-SQL},
  author  = {Yao, Zhewei and Sun, Guoheng and Borchmann, Lukasz and Nuti, Gaurav and Shen, Zheyu and Deng, Minghang and Zhai, Bohan and Zhang, Hao and Li, Ang and He, Yuxiong},
  journal = {arXiv preprint arXiv:2505.20315},
  year    = {2025}
}

@article{zhang2023sciencebenchmark,
  title={Sciencebenchmark: A complex real-world benchmark for evaluating natural language to sql systems},
  author={Zhang, Yi and Deriu, Jan and Katsogiannis-Meimarakis, George and Kosten, Catherine and Koutrika, Georgia and Stockinger, Kurt},
  journal={arXiv preprint arXiv:2306.04743},
  year={2023}
}

@article{yang2025qwen3,
  title={Qwen3 technical report},
  author={Yang, An and Li, Anfeng and Yang, Baosong and Zhang, Beichen and Hui, Binyuan and Zheng, Bo and Yu, Bowen and Gao, Chang and Huang, Chengen and Lv, Chenxu and others},
  journal={arXiv preprint arXiv:2505.09388},
  year={2025}
}

@article{zhang2025qwen3,
  title={Qwen3 embedding: Advancing text embedding and reranking through foundation models},
  author={Zhang, Yanzhao and Li, Mingxin and Long, Dingkun and Zhang, Xin and Lin, Huan and Yang, Baosong and Xie, Pengjun and Yang, An and Liu, Dayiheng and Lin, Junyang and others},
  journal={arXiv preprint arXiv:2506.05176},
  year={2025}
}

@article{sheng2024hybridflow,
  title={Hybridflow: A flexible and efficient rlhf framework},
  author={Sheng, Guangming and Zhang, Chi and Ye, Zilingfeng and Wu, Xibin and Zhang, Wang and Zhang, Ru and Peng, Yanghua and Lin, Haibin and Wu, Chuan},
  journal={arXiv preprint arXiv:2409.19256},
  year={2024}
}

@inproceedings{yu2019sparc,
  title     = {{SParC}: Cross-Domain Semantic Parsing in Context},
  author    = {Yu, Tao and Zhang, Rui and Yasunaga, Michihiro and Tan, Yi Chern and Lin, Xi Victoria and Li, Suyi and Er, Heyang and Li, Irene and Pang, Bo and Chen, Tao and Ji, Emily and Dixit, Shreya and Proctor, David and Shim, Sungrok and Kraft, Jonathan and Zhang, Vincent and Xiong, Caiming and Socher, Richard and Radev, Dragomir},
  booktitle = {Proceedings of the 57th Annual Meeting of the Association for Computational Linguistics ({ACL})},
  pages     = {4511--4523},
  year      = {2019},
  doi       = {10.18653/v1/P19-1443},
  url       = {https://aclanthology.org/P19-1443/}
}

@inproceedings{yu2019cosql,
  title     = {{CoSQL}: A Conversational Text-to-{SQL} Challenge Towards Cross-Domain Natural Language Interfaces to Databases},
  author    = {Yu, Tao and Zhang, Rui and Er, Heyang and Li, Suyi and Xue, Eric and Pang, Bo and Lin, Xi Victoria and Tan, Yi Chern and Shi, Tianze and Li, Zihan and Jiang, Youxuan and Yasunaga, Michihiro and Shim, Sungrok and Chen, Tao and Fabbri, Alexander and Li, Zifan and Chen, Luyao and Zhang, Yuwen and Dixit, Shreya and Zhang, Vincent and Xiong, Caiming and Socher, Richard and Lasecki, Walter and Radev, Dragomir},
  booktitle = {Proceedings of the 2019 Conference on Empirical Methods in Natural Language Processing and the 9th International Joint Conference on Natural Language Processing ({EMNLP-IJCNLP})},
  pages     = {1962--1979},
  year      = {2019},
  doi       = {10.18653/v1/D19-1204},
  url       = {https://aclanthology.org/D19-1204/}
}

@inproceedings{guo2018question,
  title     = {Question Generation from {SQL} Queries Improves Neural Semantic Parsing},
  author    = {Guo, Daya and Sun, Yibo and Tang, Duyu and Duan, Nan and Yin, Jian and Chi, Hong and Cao, James and Chen, Peng and Zhou, Ming},
  booktitle = {Proceedings of the 2018 Conference on Empirical Methods in Natural Language Processing ({EMNLP})},
  pages     = {1597--1607},
  year      = {2018},
  doi       = {10.18653/v1/D18-1188},
  url       = {https://aclanthology.org/D18-1188/}
}

@inproceedings{wu2021data,
  title     = {Data Augmentation with Hierarchical {SQL}-to-Question Generation for Cross-domain Text-to-{SQL} Parsing},
  author    = {Wu, Kun and Wang, Lijie and Li, Zhenghua and Zhang, Ao and Xiao, Xinyan and Wu, Hua and Zhang, Min and Wang, Haifeng},
  booktitle = {Proceedings of the 2021 Conference on Empirical Methods in Natural Language Processing ({EMNLP})},
  pages     = {8974--8983},
  year      = {2021},
  doi       = {10.18653/v1/2021.emnlp-main.707},
  url       = {https://aclanthology.org/2021.emnlp-main.707/}
}

@inproceedings{zhong2020grounded,
  title     = {Grounded Adaptation for Zero-shot Executable Semantic Parsing},
  author    = {Zhong, Victor and Lewis, Mike and Wang, Sida I. and Zettlemoyer, Luke},
  booktitle = {Proceedings of the 2020 Conference on Empirical Methods in Natural Language Processing ({EMNLP})},
  pages     = {6869--6882},
  year      = {2020},
  doi       = {10.18653/v1/2020.emnlp-main.558},
  url       = {https://aclanthology.org/2020.emnlp-main.558/}
}

@inproceedings{liu2022augmenting,
  title     = {Augmenting Multi-Turn Text-to-{SQL} Datasets with Self-Play},
  author    = {Liu, Qi and Ye, Zihuiwen and Yu, Tao and Song, Linfeng and Blunsom, Phil},
  booktitle = {Findings of the Association for Computational Linguistics: {EMNLP} 2022},
  pages     = {5608--5620},
  year      = {2022},
  doi       = {10.18653/v1/2022.findings-emnlp.411},
  url       = {https://aclanthology.org/2022.findings-emnlp.411/}
}

@inproceedings{yang2024sense,
  title     = {Synthesizing Text-to-{SQL} Data from Weak and Strong {LLM}s},
  author    = {Yang, Jiaxi and Hui, Binyuan and Yang, Min and Yang, Jian and Lin, Junyang and Zhou, Chang},
  booktitle = {Proceedings of the 62nd Annual Meeting of the Association for Computational Linguistics (Volume 1: Long Papers)},
  pages     = {7864--7875},
  year      = {2024},
  doi       = {10.18653/v1/2024.acl-long.425},
  url       = {https://aclanthology.org/2024.acl-long.425/}
}

@inproceedings{duan2025dsqg,
  title     = {{DSQG-Syn}: Synthesizing High-quality Data for Text-to-{SQL} Parsing by Domain Specific Question Generation},
  author    = {Duan, Shaoming and Wu, Youxuan and Liu, Chuanyi and Zhang, Yuhao and Wang, Zirui and Han, Peiyi and Yu, Shengyuan and Yan, Liang and Liang, Yingwei},
  booktitle = {Findings of the Association for Computational Linguistics: {NAACL} 2025},
  pages     = {2971--2989},
  year      = {2025},
  doi       = {10.18653/v1/2025.findings-naacl.162},
  url       = {https://aclanthology.org/2025.findings-naacl.162/}
}

@article{cai2025text2sqlflow,
  title   = {{Text2SQL-Flow}: A Robust {SQL}-Aware Data Augmentation Framework for Text-to-{SQL}},
  author  = {Cai, Qifeng and Liang, Hao and Xu, Chang and Xie, Tao and Zhang, Wentao and Cui, Bin},
  journal = {arXiv preprint arXiv:2511.10192},
  year    = {2025},
  doi     = {10.48550/arXiv.2511.10192},
  url     = {https://arxiv.org/abs/2511.10192}
}

@inproceedings{zhu2025sacsql,
  title     = {Learning {SQL} Like a Human: Structure-Aware Curriculum Learning for Text-to-{SQL} Generation},
  author    = {Zhu, Xiaohu and Li, Qian and Cui, Lizhen and Du, Yuntao},
  booktitle = {Findings of the Association for Computational Linguistics: {EMNLP} 2025},
  pages     = {3545--3559},
  year      = {2025},
  doi       = {10.18653/v1/2025.findings-emnlp.190},
  url       = {https://aclanthology.org/2025.findings-emnlp.190/}
}

@article{wang2018executionguided,
  title   = {Execution-Guided Neural Program Decoding},
  author  = {Wang, Chenglong and Huang, Po-Sen and Polozov, Oleksandr and Brockschmidt, Marc and Singh, Rishabh},
  journal = {arXiv preprint arXiv:1807.03100},
  year    = {2018},
  doi     = {10.48550/arXiv.1807.03100},
  url     = {https://arxiv.org/abs/1807.03100}
}

@inproceedings{he2025starsql,
  title     = {{STaR-SQL}: Self-Taught Reasoner for Text-to-{SQL}},
  author    = {He, Mingqian and Shen, Yongliang and Zhang, Wenqi and Peng, Qiuying and Wang, Jun and Lu, Weiming},
  booktitle = {Proceedings of the 63rd Annual Meeting of the Association for Computational Linguistics (Volume 1: Long Papers)},
  pages     = {24365--24375},
  year      = {2025},
  doi       = {10.18653/v1/2025.acl-long.1187},
  url       = {https://aclanthology.org/2025.acl-long.1187/}
}

@article{kulkarni2025reinforcing,
  title   = {Reinforcing Code Generation: Improving Text-to-{SQL} with Execution-Based Learning},
  author  = {Kulkarni, Atharv and Srikumar, Vivek},
  journal = {arXiv preprint arXiv:2506.06093},
  year    = {2025},
  doi     = {10.48550/arXiv.2506.06093},
  url     = {https://arxiv.org/abs/2506.06093}
}

@inproceedings{weng2025graphreward,
  title     = {{Graph-Reward-SQL}: Execution-Free Reinforcement Learning for Text-to-{SQL} via Graph Matching and Stepwise Reward},
  author    = {Weng, Han and Wu, Puzhen and Cui, Longjie and Zhan, Yi and Liu, Boyi and Song, Yuanfeng and Zeng, Dun and Yang, Yingxiang and Zhang, Qianru and Huang, Dong and Yin, Xiaoming and Sun, Yang and Chen, Xing},
  booktitle = {Findings of the Association for Computational Linguistics: {EMNLP} 2025},
  pages     = {12917--12943},
  year      = {2025},
  doi       = {10.18653/v1/2025.findings-emnlp.694},
  url       = {https://aclanthology.org/2025.findings-emnlp.694/}
}

@inproceedings{dai2025reexsql,
  title     = {{ReEx-SQL}: Reasoning with Execution-Aware Reinforcement Learning for Text-to-{SQL}},
  author    = {Dai, Yaxun and Xie, Wenxuan and Zhuang, Xialie and Yang, Tianyu and Liu, Ziyi and Yang, Haiqin and Yang, Yiying and Zhao, Yuhang and Chao, Pingfu and Jiang, Wenhao},
  booktitle = {Proceedings of the 64th Annual Meeting of the Association for Computational Linguistics (Volume 1: Long Papers)},
  pages     = {824--847},
  year      = {2026},
  doi       = {10.18653/v1/2026.acl-long.35},
  url       = {https://aclanthology.org/2026.acl-long.35/}
}

@article{hua2026sqltrail,
  title   = {{SQL-Trail}: Multi-Turn Reinforcement Learning with Interleaved Feedback for Text-to-{SQL}},
  author  = {Hua, Harper and Han, Zhen and Shen, Zhengyuan and Lee, Jeremy and Guan, Patrick and Zhu, Qi and Jeoung, Sullam and Chen, Yueyan and Bai, Yunfei and Wang, Shuai and Ioannidis, Vassilis and Rangwala, Huzefa},
  journal = {arXiv preprint arXiv:2601.17699},
  year    = {2026},
  doi     = {10.48550/arXiv.2601.17699},
  url     = {https://arxiv.org/abs/2601.17699}
}

@article{yang2024qwen2,
  title={Qwen2 technical report},
  author={Yang, An and Yang, Baosong and Hui, Binyuan and Zheng, Bo and Yu, Bowen and Zhou, Chang and Li, Chengpeng and Li, Chengyuan and Liu, Dayiheng and Huang, Fei and others},
  journal={arXiv preprint arXiv:2407.10671},
  year={2024}
}

@article{guo2024deepseek,
  title={DeepSeek-Coder: When the Large Language Model Meets Programming--The Rise of Code Intelligence},
  author={Guo, Daya and Zhu, Qihao and Yang, Dejian and Xie, Zhenda and Dong, Kai and Zhang, Wentao and Chen, Guanting and Bi, Xiao and Wu, Yu and Li, YK and others},
  journal={arXiv preprint arXiv:2401.14196},
  year={2024}
}

@inproceedings{huang2025opencoder,
  title={Opencoder: The open cookbook for top-tier code large language models},
  author={Huang, Siming and Cheng, Tianhao and Liu, Jason Klein and Xu, Weidi and Hao, Jiaran and Song, Liuyihan and Xu, Yang and Yang, Jian and Liu, Jiaheng and Zhang, Chenchen and others},
  booktitle={Proceedings of the 63rd Annual Meeting of the Association for Computational Linguistics (Volume 1: Long Papers)},
  pages={33167--33193},
  year={2025}
}

@article{mishra2024granite,
  title={Granite code models: A family of open foundation models for code intelligence},
  author={Mishra, Mayank and Stallone, Matt and Zhang, Gaoyuan and Shen, Yikang and Prasad, Aditya and Soria, Adriana Meza and Merler, Michele and Selvam, Parameswaran and Surendran, Saptha and Singh, Shivdeep and others},
  journal={arXiv preprint arXiv:2405.04324},
  year={2024}
}

\end{document}